\documentclass[iicol,pdflatex,sn-mathphys-num]{sn-jnl}
\usepackage{array}          
\usepackage{multirow}       
\usepackage{hhline}         
\usepackage{booktabs}       

\usepackage{amsmath,amssymb,amsfonts} 
\usepackage{amsthm}
\usepackage{mathrsfs}
\usepackage[title]{appendix}
\usepackage{textcomp}
\usepackage{manyfoot}
\usepackage{algorithm}
\usepackage{algorithmicx}
\usepackage{algpseudocode}
\usepackage{listings}
\usepackage{xspace}
\usepackage[table]{xcolor}

\usepackage{graphicx}
\usepackage[export]{adjustbox}
\usepackage{makecell}
\usepackage{url}
\usepackage{hyperref}
\usepackage{lineno}
\usepackage{caption}

\usepackage{tikz}
\usepackage[misc,geometry]{ifsym}
\usepackage{mathtools}
\usepackage{graphicx} 

\usepackage[table]{xcolor} 

\definecolor{bestColor}{RGB}{255, 0, 0}    
\definecolor{secondBestColor}{RGB}{0, 0, 255} 
\definecolor{thirdBestColor}{RGB}{240,0, 240}  
\definecolor{qBestColor}{RGB}{0,215, 0}  

\definecolor{secondBestColor}{RGB}{0, 0, 255} 
\definecolor{thirdBestColor}{RGB}{240,0, 240}  
\definecolor{qBestColor}{RGB}{0,215, 0}  

\definecolor{cnn_color}{RGB}{240,248,255}  
\definecolor{incompression_color}{RGB}{255,250,240}  
\definecolor{crosscompression_color}{RGB}{240,255,240}  
\definecolor{input_color}{RGB}{255,240,240}      
\definecolor{mamba_color}{RGB}{235,235,225}  
\newcommand{\pub}[1]{{\color{gray}{\tiny{[{#1}]}}}}
\definecolor{f}{HTML}{ff9999}
\definecolor{s}{HTML}{ffcc99}
\definecolor{t}{HTML}{fff8ae}

\definecolor{deepred}{RGB}{133,0,0}    
\definecolor{lightred}{RGB}{255,102,102}  
\definecolor{deepgreen}{RGB}{0,133,0}  

\definecolor{bestcrossmanupulation_color}{RGB}{255,240,240} 
\definecolor{secondcrossmanupulation_color}{RGB}{240,240,255} 
\definecolor{thirdcrossmanupulation_color}{RGB}{240,255,240} 
\definecolor{background_color}{RGB}{255,255,255}

\newcommand{\addblankpages}[1]{
    \ifnum\value{page}>1\newpage\fi
    \begingroup
    \pagestyle{empty}
    \count0=#1
    \loop
        \ifnum\count0>0
            \null
            \newpage
            \advance\count0 by -1
    \repeat
    \endgroup
}

\theoremstyle{thmstyleone}%
\theoremstyle{thmstyletwo}%

\theoremstyle{thmstylethree}%

\begin{document}

\title{SplatCo: Structure-View Collaborative Gaussian Splatting for Detail-Preserving Rendering of Large-Scale Unbounded Scenes}

\author[1]{\fnm{Haihong} \sur{Xiao}}\email{hhxiao@hfut.edu.cn}
\equalcont{These authors contributed equally to this work.}

\author[2]{\fnm{Jianan} \sur{Zou}}\email{aujazou@mail.scut.edu.cn}
\equalcont{These authors contributed equally to this work.}

\author[2]{\fnm{Yuxin} \sur{Zhou}}\email{202164690558@mail.scut.edu.cn}

\author[3]{\fnm{Ying} \sur{He}}\email{yhe@ntu.edu.sg}

\author*[2]{\fnm{Wenxiong} \sur{Kang}}\email{auwxkang@scut.edu.cn}

\affil[1]{\orgdiv{School of Computer Science and Information Engineering}, \orgname{Hefei University of Technology}, \orgaddress{ \city{Hefei} \postcode{230088}, \country{China}}}

\affil[2]{\orgdiv{School of Automation Science and Engineering}, \orgname{South China University of Technology}, \orgaddress{ \city{Guangzhou} \postcode{510641}, \country{China}}}

\affil[3]{\orgdiv{College of Computing and Data Science}, \orgname{Nanyang Technological University},\orgaddress{ \city{Singapore} \postcode{639798}, \country{Singapore}}}

\abstract{
Neural Radiance Fields (NeRFs) have achieved impressive results in novel view synthesis but are less suited for large-scale scene reconstruction due to their reliance on dense per-ray sampling, which limits scalability and efficiency. In contrast, 3D Gaussian Splatting (3DGS) offers a more efficient alternative to computationally intensive volume rendering, enabling faster training and real-time rendering. Although recent efforts have extended 3DGS to large-scale settings, these methods often struggle to balance global structural coherence with local detail fidelity. Crucially, they also suffer from Gaussian redundancy due to a lack of effective geometric constraints, which further leads to rendering artifacts. To address these challenges, we present SplatCo, a structure–view collaborative Gaussian splatting framework for high-fidelity rendering of complex outdoor scenes. SplatCo builds upon three novel components: (1) a Cross-Structure Collaboration Module (CSCM) that combines global tri-plane representations, which capture coarse scene layouts, with local context grid features that represent fine details. This fusion is achieved through the proposed hierarchical compensation mechanism, ensuring both global spatial awareness and local detail preservation; (2) a Cross-View Pruning Mechanism (CVPM) that prunes overfitted or inaccurate Gaussians based on structural consistency, thereby improving storage efficiency while avoiding Gaussian rendering artifacts; (3) a Structure–View Co-learning (SVC) Module that aggregates structural gradients with view gradients, redirecting the Gaussian geometric and appearance attribute optimization more robustly guided by additional structural gradient flow. By combining these key components, SplatCo effectively achieves high-fidelity rendering for large-scale scenes. Comprehensive evaluations on 13 diverse large-scale scenes, including Mill19, MatrixCity, Tanks \& Temples, WHU, and custom aerial captures, demonstrate that SplatCo establishes a new benchmark for high-fidelity rendering of large-scale unbounded scenes. Code and project page are available at \url{https://splatco-tech.github.io/}.
}

\keywords{3DGS, real-time rendering, large-scale unbounded scenes, detail preservation, structure-view collaborative learning}

\maketitle

	
    \section{Introduction}
	\label{sec:intro}
	Neural scene representations have significantly advanced novel view synthesis and 3D reconstruction. Among them, Neural Radiance Fields (NeRFs) \cite{relate_nerf} achieve impressive photorealistic rendering of objects and small-scale scenes from multi‑view captures, supporting applications in augmented/virtual reality \cite{intro_vr}, autonomous driving \cite{intro_driving}, digital twins \cite{intro_twin}, telepresence, and cultural heritage preservation \cite{intro_culture}. 

Applying NeRF to large-scale outdoor and urban environments \cite{exp_meganerf,exp_switchnerf} remains challenging due to its inherent computational and representational limitations. A key difficulty lies in performing differentiable operations over densely sampled 3D points across entire scenes. Even simple operations—such as evaluating a small Multi-Layer Perceptron (MLP) per point—become extremely memory intensive, as all intermediate values must be stored for backpropagation. To overcome these limitations, 3D Gaussian Splatting (3DGS) \cite{relate_3dgs} has emerged as an efficient alternative. It represents a scene as a sparse collection of anisotropic 3D Gaussians whose positions, scales, orientations, and colors are optimized directly from images. These Gaussians are rendered using a tile-based GPU splatting pipeline, enabling high-quality, real‑time rendering.

While 3DGS overcomes the rendering efficiency limitations of NeRFs, its extension to large-scale outdoor scenes ~\cite{relate_new_lu2024scaffold,mvgs,liu2024citygaussian} faces three core challenges: global structural coherence, local detail fidelity, and robust cross-view consistency. Current methods address these issues in isolation. Techniques such as Scaffold-GS utilize anchor-based context-grids \cite{method_contextgs,method_hash_grid} to capture fine local details but lack global awareness, while tri-plane representations \cite{method_igs,method_trigs} model the scene's global features effectively yet lose high-frequency detail information. To balance efficiency and memory, Octree-GS \cite{ren2024octree} adaptively selects appropriate levels from a multi-scale set of Gaussian primitives to ensure consistent rendering performance. However, it fundamentally lacks an effective mechanism to leverage cross-view consistency for accurately identifying and pruning geometrically inaccurate or redundant Gaussians.

Recent promising methods have attempted to address the limitations of conventional 3DGS from multiple perspectives. For example, MVGS \cite{mvgs} introduces a multi-view regulated training strategy to address the limitations of single-view supervision, VastGaussian~\cite{lin2024vastgaussian} employs a partition-and-merge strategy to accelerate training convergence, and CityGS-X \cite{citygs-x} proposes a parallelized hybrid hierarchical 3D representation. Nevertheless, these approaches still face the following critical challenges: (1) the absence of a scalable feature encoding scheme for large-scale scenes; (2) the lack of interaction between partitioned blocks leads to inconsistencies at boundaries due to insufficient multi-view constraints; and (3) high memory overhead from maintaining multiple representations, often requiring multiple GPUs and hindering real-time deployment. In practice, however, most users lack access to multi-GPU setups. This raises the question: can we enable large-scale 3DGS training on a single consumer-grade GPU? Therefore, a computationally economical and accessible unified framework that jointly optimizes global structure, local details, and multi-view consistency remains an open problem—and is the primary goal of our work.

\begin{figure*}[!t]
\centering
\includegraphics[width=1.0\textwidth]{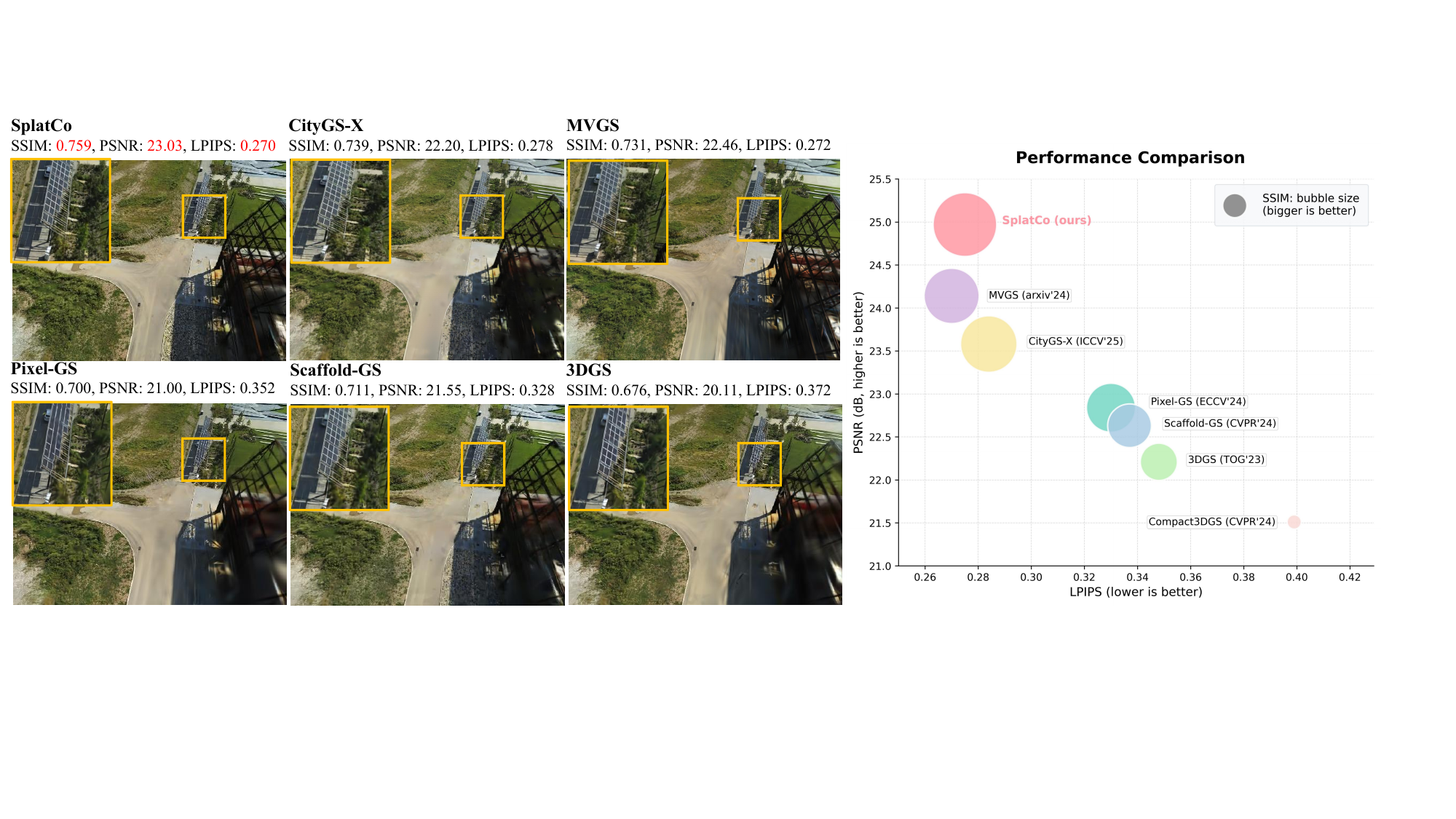}
\caption{Comparison of 3DGS-based methods on two scenes (Rubble and Building) from the Mill-19 Dataset. Yellow squares highlight regions with noticeable visual differences. SplatCo produces more realistic and detailed renderings, recovering sharper texture patterns compared to state-of-the-art methods. The chart on the right presents quantitative results across three standard rendering metrics: PSNR, SSIM and LPIPS.
}
\label{fig_cover}
\end{figure*}

In this work, we propose Structure–View Collaborative Gaussian Splatting (SplatCo), for high-fidelity rendering of large-scale outdoor scenes. Unlike prior approaches that primarily focus on supervising view consistency, SplatCo integrates both Structural-View Co-learning (SVC) to achieve robust optimization of Gaussian position and appearance attributes. To robustly extract geometric information from large-scale scenes, we incorporate multi-scale contextual information into a multi-scale tri-plane backbone using a three-level compensation mechanism, achieving a coarse-to-fine cascading optimization. We also introduce a Cross-View Pruning Mechanism (CVPM) that prunes overfitted or inaccurate Gaussians based on structural consistency, thereby improving storage efficiency while avoiding Gaussian rendering artifacts. By combining these components, SplatCo achieves robust fine-detail rendering across complex, large-scale environments, as shown in Fig.~\ref{fig_cover}. SplatCo demonstrates its consistent effectiveness in high-fidelity rendering of unbounded outdoor environments compared to state-of-the-art Gaussian-based methods, and we validate this across 13 large-scale outdoor scenes, including four public benchmarks (Mill-19, MatrixCity, Tanks \& Temples, WHU) and five custom aerial captures.

Our main contributions are as follows: \begin{itemize} 
\item We propose a structure–view collaborative Gaussian splatting framework to achieve high-fidelity rendering for large-scale scenes, enabling robust optimization of Gaussian geometry and appearance attributes through aggregating structural gradients with view gradients.

\item We develop a cross-structure collaboration module that combines global tri-plane features with local context grids through the proposed hierarchical compensation mechanism, enabling both fine detail preservation and global spatial awareness.

\item We design a cross-view pruning mechanism that prunes overfitted or inaccurate Gaussians based on structural consistency, thereby improving storage efficiency while avoiding Gaussian rendering artifacts.

\item We conduct extensive experiments on diverse large-scale outdoor scenes with varying structural forms, showing that SplatCo achieves state-of-the-art performance in detail-preserving novel view synthesis.

\end{itemize}
	\section{Related Work}
	\label{sec:review}
	\subsection{Multi-view 3D Reconstruction}
3D reconstruction has long been a central problem in computer vision, with many approaches relying on multi-view image inputs to infer scene geometry. Structure-from-Motion (SfM)~\cite{relate_new_cui2017hsfm,relate_new_schonberger2016structure} approaches rely on handcrafted features, such as SIFT~\cite{relate_new_zheng2017sift}), and bundle adjustment~\cite{relate_new_agarwal2010bundle} to recover sparse scene structures. Multi-View Stereo (MVS)~\cite{relate_tra_1,relate_tra_2} methods extend these techniques by estimating dense depth maps across multiple views, thereby producing detailed point clouds. These raw point clouds can be fed into point orientation methods, such as~\cite{Hou2022iPSR,Xu2023GCNO,Liu2024DWG,Lin2024WNNC}, to compute globally consistent orientations. Meshes can be constructed from the oriented point clouds using surface reconstruction techniques such as Poisson surface reconstruction~\cite{Kazhdan2006,Kazhdan2013}.

Learning-based MVS methods~\cite{relate_mvsnet,relate_mvster,relate_rmvsnet,relate_casmvsnet,relate_transmvsnet,relate_unimvsnet} have demonstrated significant improvements in reconstruction accuracy and robustness by adopting a standardized four-stage pipeline: (1) feature extraction, (2) differentiable homography, (3) cost volume regularization, and (4) depth map regression. Variants have improved speed and scalability using recurrent networks~\cite{relate_rmvsnet}, coarse-to-fine cascades~\cite{relate_casmvsnet}, and Transformer backbones~\cite{relate_transmvsnet}. More compact and flexible depth representations have also been explored~\cite{relate_unimvsnet}. However, cost volumes are memory-intensive and resolution-limited, making these methods less practical for real-time or outdoor-scale reconstruction.

Recently, feature-matching-based approaches, such as DUST3R~\cite{relate_dust3r} and its variants~\cite{relate_must3r,relate_fast3r,relate_d2ust3r}, further reduce reliance on depth volumes by directly matching features across images. These methods offer a promising alternative for scalable reconstruction.
\subsection{Novel View Synthesis}
Novel view synthesis has become a central task in visual computing, with broad relevance to virtual ad augmented reality, robotics, and and autonomous systems. Early work used explicit representations, such as meshes~\cite{relate_new_mesh_liu2019liquid,realet_new_mesh_guo2023vmesh}, voxels ~\cite{relate_new_voxel_sun2023vgos,relate_new_voxel_schwarz2022voxgraf}, and point clouds~\cite{relate_new_pointcloud_xiao2023distinguishing}, to model scenes. However, these methods struggled to reproduce fine visual details.

NeRFs~\cite{relate_nerf} introduce a view-consistent rendering framework based on implicit volumetric representations learned from 2D images. While NeRF methods produce realistic results in controlled scenes, their sampling inefficiency and reliance on large MLPs lead to slow training and inference. These issues are especially pronounced in outdoor or large-scale environments. To improve performance, various acceleration methods have been proposed~\cite{relate_new_korhonen2024efficient,relate_new_li2023nerfacc}, along with anti-aliasing~\cite{relate_mipnerf,relate_new_hu2023multiscale}, motion deblurring~\cite{relate_deblur_nerf,relate_new_wang2024mp}, and support for dynamic content~\cite{relate_dnerf}. However, the fundamental design of NeRFs still poses scalability and efficiency challenges.

3D Gaussian Splatting~\cite{relate_3dgs} offers a more efficient rendering pipeline by replacing volume sampling with rasterization and representing the scene as a set of Gaussian primitives. This design enables real-time rendering with competitive visual quality and has become the basis for a growing body of recent work.

\subsection{3D Gaussian Splatting for Large-Scale Scenes}
3DGS has been widely adopted for modeling objects and scenes due to its speed and rendering quality. Readers may refer to recent surveys for comprehensive overviews of 3DGS and its applications~\cite{chen2025survey3dgaussiansplatting,fei3dgssurvey,gaosurvey}. While early applications focused on compact scenes, such as individual objects~\cite{relate_new_pryadilshchikov2024t,relate_new_sun2024f}, humans~\cite{relate_new_sakamiya2024avatarperfect,relate_new_chen2025thgs,relate_new_qiu2024anigs}, and indoor environments~\cite{relate_gsdf,relate_gsrec,relate_gsroom}, adapting 3DGS to large-scale outdoor scenes remains a challenge. 

To bridge this gap, recent works propose scalable variants of 3DGS. Scaffold-GS~\cite{relate_new_lu2024scaffold} introduces a hierarchical, region-aware 3D representation via anchor points. DOGS~\cite{chen2024dogs} significantly improves training efficiency through a distributed paradigm that combines spatial partitioning with the ADMM solver. VastGaussian~\cite{lin2024vastgaussian} and Hierarchy-GS~\cite{kerbl2024hierarchical} explore alternative partitioning strategies, each offering distinct architectural advantages. More recent developments include Octree-GS~\cite{ren2024octree}, which organizes Gaussian primitives in an octree structure to support adaptive multi-scale rendering, and CityGaussian~\cite{liu2024citygaussian}, which incorporates level-of-detail techniques to optimize both training and rendering in urban-scale environments. Notably, MVGS~\cite{mvgs} introduces a multi-view regulated training strategy to address the limitations of single-view supervision in the original 3DGS, thereby offering partial inspiration for our work.

Despite these advances, two fundamental limitations persist. First, rigid partitioning of space often breaks continuity of semantic structures, such as roads, facades or vegetation. Second, the lack of structural coordination and view-collaboration mechanisms hampers the provision of stable geometric guidance necessary for large-scale scene rendering. These challenges often lead to visual artifacts, such as ghosting, texture flickering, and inconsistent shading. 
\begin{figure*}[!t]
\centering
\includegraphics[width=1.0\textwidth]{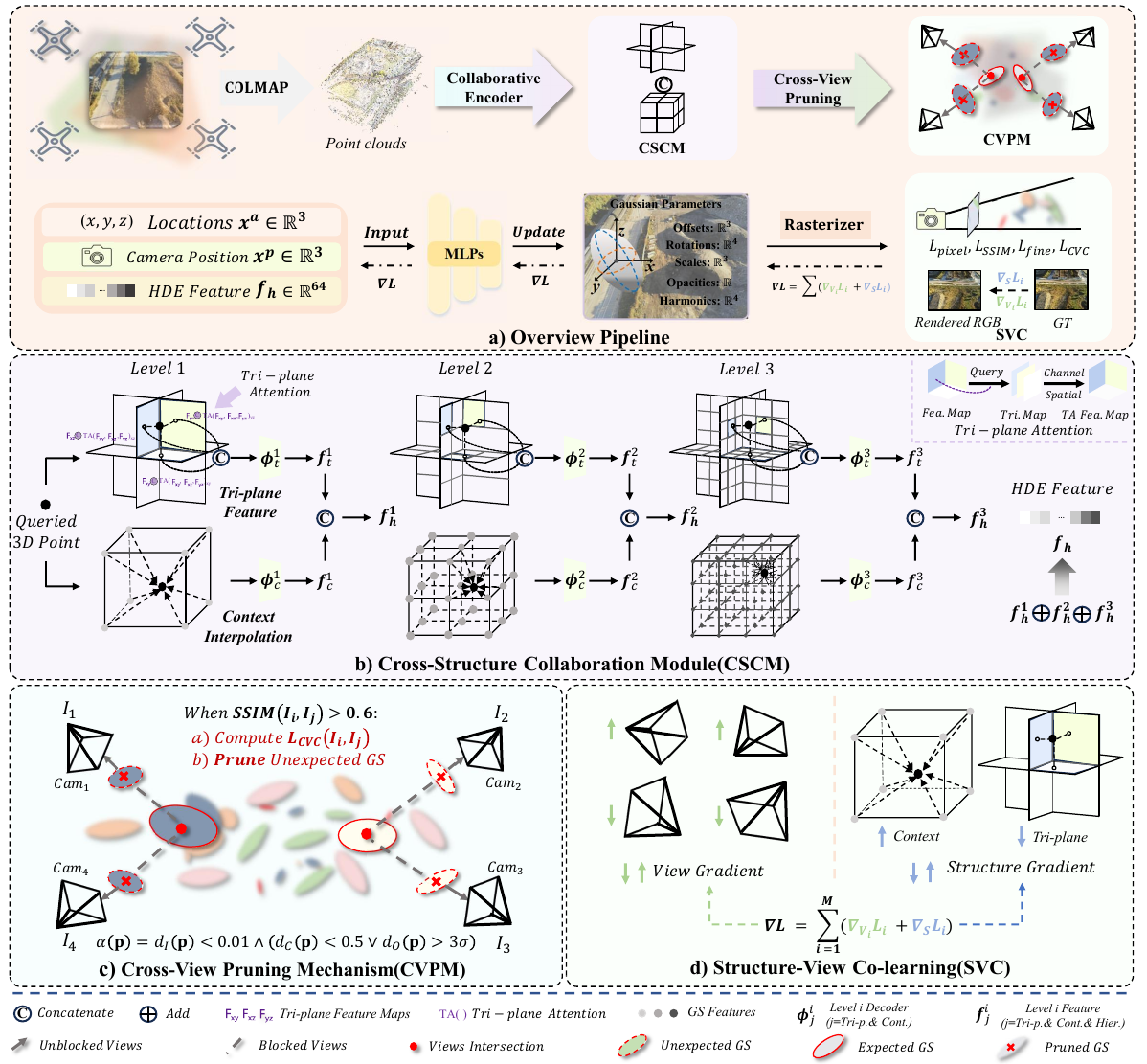}
\caption{\textbf{Overview}. we first introduce the CSCM, which integrates multi-scale tri-planes and context-aware grids to generate Hierarchical Dense Embedded features ($f_h$), simultaneously ensuring global coherence and local detail preservation. Subsequently, the CVPM maintains integrity by employing a Cross-View Consistency Loss ($L_{CVC}$) and a geometric consistency-based pruning mechanism to reduce redundancy. Finally, the SVC module combines structural and view-dependent gradients for robust joint optimization.}
\label{Fig1}
\end{figure*}
    
	\section{Method}
	\label{sec:method}
	\subsection{Motivations \& Overview}
\label{subsec:overview}
We propose a 3DGS framework tailored for large-scale scenes to achieve high-fidelity scene rendering. As illustrated in Fig.~\ref{Fig1}, our framework addresses two key limitations of existing approaches:

First, current 3DGS methods \cite{method_contextgs, method_igs, method_trigs} either employ tri-plane structures or use context grid interpolation for spatial modeling. Both technologies have inherent limitations: Tri-plane representations often suffer from detail loss due to fixed-axis-aligned partitioning, while context interpolation lacks global spatial awareness. These issues result in reduced structural expressiveness and visual artifacts.

Second, traditional single-view optimization schemes \cite{relate_3dgs} in 3DGS exhibit two major drawbacks: 1) susceptibility to overfitting to individual views, and 2) the accumulation of redundant Gaussians near camera paths, leading to inefficient storage and training, especially in large-scale scenes.

To address the aforementioned limitations, we first propose a CSCM that fuses multi-scale tri-plane features with context-aware grid features, enabling the simultaneous preservation of global spatial coherence and local geometric details. Second, we design a CVPM, which comprises: 1) a cross-view consistency loss designed to enforce structural coherence across diverse perspectives, and 2) an adaptive pruning strategy governed by geometric consistency to reduce the redundancy of Gaussians. Third, we introduce a SVC module to synergize structural and view gradients, thereby enabling better optimization of Gaussian positions and appearance attributes.
\subsection{Preliminary for 3D Gaussian Splatting}
3DGS \cite{relate_3dgs} proposes a new paradigm for scene representation, utilizing anisotropic 3D Gaussians as its core primitives. This method effectively integrates the differentiability inherent in volumetric representations with the computational efficiency of tile-based rasterization, achieving rapid rendering while preserving the capacity for gradient-based optimization.

3DGS begins by deriving a sparse point cloud from multi-view images using COLMAP, exemplified by tools such as Structure-from-Motion (SfM) techniques. Each Gaussian is parameterized by a mean position $\mu \in \mathbb{R}^3$ and a covariance matrix $\Sigma \in \mathbb{R}^{3 \times 3}$, which jointly describe its anisotropic spatial distribution. The Gaussian function is mathematically defined as:
\begin{equation}
\label{equ1}
G(x) = \exp\left(-\frac{1}{2} (x - \mu)^\top \Sigma^{-1} (x - \mu)\right),
\end{equation}
where $x \in \mathbb{R}^3$ denotes a point in the 3D space. To ensure positive semi-definiteness of the covariance matrix, it is decomposed as:
\begin{equation}
\label{equ2}
\Sigma = R S S^\top R^\top,
\end{equation}
where $S \in \mathbb{R}^{3 \times 3}$ is a diagonal matrix encoding scaling factors along principal axes, and $R \in \mathbb{R}^{3 \times 3}$ is a rotation matrix aligning the Gaussian with the scene geometry.

For appearance modeling, each Gaussian is augmented with an opacity parameter $\alpha \in [0, 1]$ to control transparency and spherical harmonic coefficients ($SHs$) to capture view-dependent color effects, represented as $c$. During rendering, the opacity $\alpha$ modulates the Gaussian density $G(x)$, enabling precise regulation of translucency and supporting smooth blending of overlapping primitives.

The rendering process in 3DGS employs tile-based rasterization to optimize computational efficiency, circumventing the intensive demands of ray marching. Specifically, 3D Gaussians $G(x)$ are projected onto the image plane, producing 2D Gaussian distributions $G'(x')$. A specialized tile-based rasterizer efficiently sorts and manages these projections by depth, applying $\alpha$-blending to composite their contributions into a final pixel color, expressed as:
\begin{equation}
\label{equ3}
C(x') = \sum_{i \in N} c_i \sigma_i \prod_{j=1}^{i-1} (1 - \sigma_j), \quad \sigma_i = \alpha_i G'_i(x'),
\end{equation}
where $x' \in \mathbb{R}^2$ represents a pixel coordinate, and $N$ denotes the set of depth-ordered 2D Gaussians contributing to that pixel. The differentiable nature of this rasterization pipeline facilitates end-to-end optimization of all Gaussian attributes—position, covariance, opacity, and color—using gradient-based methods, offering a compelling alternative to traditional volumetric rendering approaches.

\subsection{Cross-Structure Collaboration Module}
The tri-plane and context grid structures are two powerful representations for encoding 3D point cloud features. Both methods utilize interpolation to extract features at any arbitrary point within the 3D space. The former achieves this by applying interpolation across the entire point cloud feature map, while the latter accomplishes it through local resolution grids. However, both of these structures have their respective limitations: \emph{the tri-plane structure emphasizes global features while neglecting the local relationships between points in the point cloud, and the context grid structure focuses on interpolation within resolution grids, making it challenging to capture the spatial correlations of the entire point cloud.}

Therefore, we propose the CSCM, which integrates the global perception of the tri-plane structure with the local sensitivity of the context grid structure. This approach facilitates more accurate feature extraction for each Gaussian point within the overall point cloud, significantly improving the quality of rendering details in subsequent novel view synthesis. Inspired by Feature Pyramid Networks (FPN) \cite{method_fpn}, we divide the overall training process into three levels. During training, features are progressively compensated from coarse to fine, aiming to reduce feature misalignment caused by the size of the tri-plane feature map and grid resolution. Specifically, the CSCM consists of three components: \textbf{tri-plane feature acquisition}, \textbf{context interpolation}, and \textbf{cross-structure hierarchical compensation}.

\textbf{Tri-plane Feature Acquisition.}
Tri-Plane is a static 3D instance of the $k$-planes factorization \cite{method_kplanes}, which represents the 3D space using three $m$-channel feature plane maps $\mathbf{F} \in \mathbb{R}^{w \times h \times m}$, each corresponding to one of the axis-aligned planes: $xy$, $xz$, and $yz$, with spatial resolution $w \times h$. To obtain fine-grained tri-plane features for a queried 3D point $\mathbf{p}$, we first apply a tri-plane attention module to generate attention-enhanced tri-plane features $\mathbf{F}^{TA}$. This module captures relative features across all three planes from the global tri-plane context, thereby enhancing the global perception capability compared to traditional tri-plane representations that focus solely on individual planes. The formulation is as follows:
\begin{equation}
\label{equ4}
\mathbf{F}^{TA}_i = \text{TA}(\mathbf{F}_i),
\end{equation}
\begin{equation}
\label{equ5}
\text{TA}(\mathbf{F}_i) = \rho_i(\text{SA}(\text{CA}(||_{j \in \{xy, xz, yz\}}(\mathbf{F}_j))),
\end{equation}
where $i$ denotes the axis-aligned planes: $xy$, $xz$, and $yz$, and $\text{TA}(\cdot)$ represents the tri-plane attention operation. The operator $||_{j \in \{xy, xz, yz\}}(\cdot)$ indicates concatenating feature planes along the channel dimension in the order of $xy$, $xz$, and $yz$. $\text{SA}(\cdot)$ and $\text{CA}(\cdot)$ denote the spatial attention and channel attention operations \cite{method_cbam} applied to the feature maps, respectively. $\rho_i(\cdot)$ corresponds to the operation of extracting the tri-plane attention feature map for the dimension indexed by $i$ from the previously concatenated feature maps.

We then project the given 3D point $\mathbf{p}$ onto these planes and interpolate the features on the corresponding grids. The interpolated features are then concatenated to form the feature representation for this point. Finally, the decoder $\phi_t$ is used to obtain the tri-plane feature ${f}_t$ for the point:
\begin{equation}
\label{equ6}
{f}_t = \phi_t(||_{i \in \{xy, xz, yz\}}(\psi(\mathbf{F}_i, \pi_i(\mathbf{p})), \psi(\mathbf{F}^{TA}_i, \pi_i(\mathbf{p}))),
\end{equation}
where $\pi_i(\cdot)$ is the projection function, and $\psi(\cdot)$ denotes bi-linear interpolation on the feature map. $\phi_t(\cdot)$ represents the tri-plane feature decoder. 

Finally, We concatenate the interpolated original features from both the original and attention-enhanced feature maps of the same orientation. The features from different tri-plane orientations are sequentially concatenated along the channel dimension. The concatenated features are passed through the decoder to obtain the tri-plane feature $f_t$.

\textbf{Context Interpolation.} In addition to leveraging tri-plane features for global scene perception, we also exploit the local intrinsic relationships between the three-dimensional Gaussians to enrich the scene representation. To achieve this, we employ grid-based structures to obtain contextual interpolation features for each Gaussian point, which are typically compact, well-organized, and capable of establishing spatial connections among the Gaussians.

Specifically, for the current query Gaussian point \( p \), we first create a grid structure for the current point cloud based on the given resolution \( \alpha \). As shown in Fig. \ref{Fig1}(b), the vertices of the cube with blue contours represent the context Gaussians for interpolation of the query Gaussian. Then, based on the distance between the query Gaussian point \( p \) and the vertices of the corresponding grid, we use the features \( f_g \) of the vertex Gaussians to perform a weighted aggregation. Finally, the decoder $\phi_c$ is used to decode and obtain the contextual feature \( f_c \) for the current point:
\begin{equation}
\label{equ7}
f_c = \phi_c\left(\frac{\sum_{v \in G} {w_v \times f_g}}{\sum_{v \in G} w_v}\right),
\end{equation}
\begin{equation}
\label{equ8}
w_v = \prod_{\text{dim} \in \{x, y, z\}} \left( 1-\{ \mathbf{p}_{\text{dim}} - v_{\text{dim}} \} \right),
\end{equation}
\begin{equation}
\label{equ9}
f_g = [f_p, p, O, s ],
\end{equation}
where \( G \) is the resolution grid that the query point \( \mathbf{p} \) falls on, and \( v \) represents the vertices of \( G \). The notation \( \{\cdot\} \) is used to compute the normalized absolute distance between the coordinates of two points along each axis, while \( [\cdot] \) denotes the concatenation operation. \( f_p \) refers to the inherent features of the query point, \( O \) is the offset of the neural Gaussian center associated with the query point, and \( s \) denotes the scales of these Gaussians.

\textbf{Cross-Structure Hierarchical Compensation.} 
Once we have separately obtained the global features \( f_t \) of the query point \( p \) from the tri-plane structure and the local features \( f_c \) from the contextual grid interpolation, we propose a cross-structure compensation mechanism to fully fuse the advantages of both types of features. This mechanism results in the refined Hierarchical Dense Embedded (HDE) feature \( f_h \), which captures both global and local point cloud relationships:
\begin{equation}
\label{equ10}
f_h = [f_t, f_c ].
\end{equation}
Additionally, considering that the refinement of scene representation by both structural features is limited to some extent by the resolution of the feature maps and the grid resolution, we introduce a hierarchical compensation mechanism based on the cross-structure refined features. We divide the overall training process into three levels, progressively increasing the resolution of the tri-plane feature maps and utilizing finer contextual grids as the training progresses. The refined HDE features \( f_h \) from each level are then merged:
\begin{equation}
\label{equ10}
f_h = \sum_{i \in L}{f_h^i}.
\end{equation}
Here, \( L \) denotes the number of levels, and \( f_h^i \) represents the cross-structure refined features at the \( i \)-th level. Thus, \( f_h \) provides the ability to capture fine scene details for subsequent rendering.

\subsection{Cross-View Pruning Mechanism}
Although we have previously derived cross-structure collaborative HDE features \( f_h \) that effectively capture the Gaussian point geometric structure information, the inherent reliance on single-view image-based optimization in the original 3DGS method exacerbates the overfitting of Gaussians to the training viewpoint. 

Encouragingly, MVGS~\cite{mvgs} utilizes multi-view information for Gaussian densification, but this approach inevitably introduces numerous redundancies and escalates computational overhead. Since these Gaussians tend to overfit the current training viewpoint by clustering extremely close to the camera plane, they lack geometric validity for novel viewpoints despite minimizing training error. Moreover, this issue is further exacerbated in large-scale scenes where the limited overlap between viewpoints provides insufficient geometric constraints. To mitigate this issue, we introduce the CVPM, which incorporates the formulation of a Cross-View Consistency Loss and a subsequent Pruning Procedure to ensure the validity of stored Gaussians, thereby enhancing their geometric representativeness while simultaneously avoiding Gaussian rendering artifacts.

\textbf{Cross-View Consistent Loss.} To reinforce structural continuity across the scene and mitigate Gaussian overfitting, following ~\cite{mvgs}, we first select adjacent view pairs from the set of \( M \) views. The selection is governed by the Structural Similarity Index Measure (SSIM), where we retain pairs exhibiting an SSIM value greater than 0.6 to ensure sufficient geometric overlap. Subsequently, we compute the Cross-View Consistent Loss (CVCL), \( L_{\text{CVC}} \), which penalizes discrepancies between the residuals of paired views, weighted by their structural similarity:
\begin{equation}
\label{equ14}
L_{\text{CVC}} = \text{SSIM}(I_i^{gt}, I_j^{gt}) \cdot L_1 \left( (I_i^{gt} - I_i), (I_j^{gt} - I_j) \right),
\end{equation}
where \( I^{gt} \) and \( I \) denote the ground truth and rendered images, respectively. This loss enforces that regions with high structural similarity in ground truth should exhibit consistent error patterns in reconstruction, thereby encouraging global coherence.

\textbf{Pruning Procedure.} Building upon the consistency constraints, we further implement a geometric pruning strategy to reduce Gaussian redundancies. We define a binary mask \( M_f \) to identify and remove overfitted Gaussians:
\begin{equation}
\label{equ15}
M_f(\mathbf{p}) = 
\begin{cases} 
1, & \text{if} \,\, \alpha(\mathbf{p}) \text{ is True}, \\
0, & \text{otherwise},
\end{cases}
\end{equation}
where the pruning condition \( \alpha(\mathbf{p}) \) is determined by a combination of ray intersection and geometric plausibility criteria:
\begin{equation}
\label{equ16}
\alpha(\mathbf{p}) = \left( d_I(\mathbf{p}) < 0.01 \right) \land \left( d_C(\mathbf{p}) < 0.5 \lor d_O(\mathbf{p}) > 3\sigma \right).
\end{equation}
Here, \( M_f(\mathbf{p}) = 1 \) signifies that a Gaussian at position \( \mathbf{p} \) is classified as an artifact and subsequently pruned. The term \( d_I(\mathbf{p}) \) represents the perpendicular distance from the Gaussian to the camera ray; a value \( d_I < 0.01 \) implies the point lies essentially on the ray path. To filter out structurally invalid points, we evaluate two additional metrics: \( d_C(\mathbf{p}) \), the distance to the camera center, and \( d_O(\mathbf{p}) \), the distance to the point cloud centroid. A Gaussian is deemed a visual overfitting artifact that neglects structural consistency if it is either anomalously close to the camera (\( d_C < 0.5 \)) or constitutes a statistical outlier (\( d_O > 3\sigma \), where \( \sigma \) is the spatial variance of the point cloud). Through this mechanism, we refine the Gaussian distribution during training, effectively curbing the risk of overfitting, while simultaneously reducing memory storage.

\subsection{Structure-View Co-learning}
Current 3DGS approaches universally employ view-based loss functions to drive the optimization of Gaussian parameters via gradient descent. The standard formulation for the view-based gradient is expressed as:
\begin{equation}
    \nabla \mathcal{L} = \sum_{i=1}^{M} \nabla_{\mathcal{V}_i} \mathcal{L}_i,
    \label{eq:view_gradient}
\end{equation}
where \( M \) denotes the number of training views, and \( \nabla_{\mathcal{V}_i} \mathcal{L}_i \) represents the gradient derived from the \( i \)-th view.

However, in the context of large-scale scenes, this paradigm faces significant challenges. \emph{Characterized by extensive spatial coverage and sparse overlap between adjacent viewing sequences, view-dependent signals alone are insufficient to explicitly capture the holistic scene structure. Furthermore, approaches relying exclusively on view-based gradient backpropagation, whether from single\cite{relate_3dgs} or multiple perspectives\cite{mvgs}, neglect this intrinsic structural information. Consequently, this renders the reconstruction highly susceptible to overfitting, typically manifesting as severe visual artifacts and blurred geometric structures}.

To address these limitations overlooked by prior work, we naturally propose the SVC module. Moving beyond reliance solely on view-based gradients, our approach also leverages the proposed hierarchical context and multi-resolution tri-plane modules to extract structural gradient information for each viewpoint. The total gradient for joint optimization is formulated as:
\begin{equation}
    \nabla \mathcal{L} = \sum_{i=1}^{M} (\nabla_{\mathcal{V}_i} \mathcal{L}_i + \nabla_{\mathcal{S}} \mathcal{L}_i),
    \label{eq:total_gradient}
\end{equation}
where \(\nabla_{\mathcal{S}} \mathcal{L}_i\) represents the structural gradient derived from our modules. By aggregating these intrinsic scene structural gradients with view gradients, the SVC module effectively optimizes the geometric position and appearance attributes of the Gaussians. It redirects the update of Gaussian parameters from a trajectory highly susceptible to overfitting towards one that is robustly constrained by structural guidance. This paradigm shift holds profound significance for advancing both the foundational 3DGS architecture and robust Novel View Synthesis in large-scale scenarios.

\subsection{Update Gaussian Attributes $\&$ Rasterization}
To fully leverage the features previously acquired, we also employ the method of Neural Gaussians\cite{relate_scaffoldgs} obtained earlier. Specifically, we integrate Gaussian positions, camera poses, and hierarchical features through multiple MLPs to update Gaussian attributes. For each visible anchor, \( k \) Gaussians are generated and their respective attributes are computed. The position of the Gaussian primitives corresponding to an anchor at \( x^a \) is derived as follows:

\begin{equation}
\label{equ18}
\{\mu_0, \ldots, \mu_{k-1}\} = x^a + \{O_0, \ldots, O_{k-1}\} \cdot l^a,
\end{equation}
where \( \{O_0, O_1, \ldots, O_{k-1}\} \in \mathbb{R}^{k \times 3} \) are learnable offsets, and \( l^a \) is a learnable scaling factor. The attributes of the Gaussians are obtained using MLPs:

\begin{equation}
\label{equ18}
\{\mathcal{A}_0, \ldots, \mathcal{A}_{k-1}\} = \text{MLPs}([x^a, x^p, f_h]),
\end{equation}
where \( \mathcal{A}_i \) represents the parameters of the \( k \) Gaussian primitives for each anchor, and \( [\cdot] \) denotes the concatenation operation.

After determining the parameters for each Gaussian, we use rasterization methods as outlined in 3DGS\cite{relate_3dgs} to generate rendered images from the respective viewpoints. The network parameters are then optimized according to the loss functions.

	\section{Loss Function}
	\label{sec:loss}
	In summary, our loss function consists of three components: 1) rendering pixel loss \( L_\text{pixel} \), 2) Structural Similarity Index Measure (SSIM) loss \( L_\text{SSIM} \) \cite{exp_sim}, 3) fine-grained Gaussian regularization loss \( L_\text{fine} \), 4) cross-view consistent loss \( L_{\text{CVC}} \).

\subsection{Rendering Loss}
The rendering loss measures pixel-level color differences between the training and rendered images using the \( L_1 \) norm. Specifically, the \( L_\text{pixel} \) is defined as:
\begin{equation}
\label{equ19}
L_\text{pixel} = \frac{1}{N} \sum_{i=1}^N |I_i^{gt} - I_i|,
\end{equation}
where \( I_{\text{rendered}} \) denotes the rendered image, \( I_{\text{gt}} \) represents the ground truth, and \( N \) is the total number of pixels in the image.

\subsection{Structural Similarity Index Measure Loss}
SSIM is employed to assess the structural similarity between two images, evaluating differences in brightness, contrast, and structure. The loss function is formulated as:

\begin{equation}
\label{equ20}
L_\text{SSIM} = 1 - \text{SSIM}( I_i^{gt}, I_i),
\end{equation}
where the \( \text{SSIM}(\cdot) \) function computes the Structural Similarity Index between the rendered image and the ground truth.

\subsection{Fine-Grained Gaussian Regularization Loss}
To improve the ability of Gaussian primitives to capture finer scene details, we introduce a regularization loss that minimizes the volume of Gaussian ellipsoids. The loss is defined as follows:

\begin{equation}
\label{equ22}
L_\text{fine} = \sum_{i=1}^{N_g} \prod(s_i),
\end{equation}
where \( N_g \) represents the number of neural Gaussians in the scene, and \( \prod(\cdot) \) indicates the product of the elements in a vector. In this case, it applies to the scale values \( s_i \) of each Gaussian.

\subsection{Overall Loss}
The total loss is a weighted sum of the individual loss terms:
\begin{equation}
\mathcal{L} = \ (1 - \lambda_1) L_\text{pixel} + \lambda_1 L_\text{SSIM} 
              + \lambda_2 L_\text{fine} + \lambda_3 L_\text{CVC}
\end{equation}
where \( \lambda_1 \), \( \lambda_2 \) and \( \lambda_3 \) are the weightsto control the relative importance of each loss term. In our implementation, we empirically set the weights as: $\lambda_1=0.2$, $\lambda_2=0.01$ and $\lambda_3=0.05$.
	\section{Experiments}
	\label{sec:experiments}
	\begin{table*}[!t]
\caption{Quantitative Comparison on Mill-19 and MatrixCity datasets. The '-' symbol represents Mega-NeRF \cite{exp_meganerf} and Switch-NeRF \cite{exp_switchnerf} were not evaluated on MatrixCity. The best, second-best, and third-best results are marked in red, orange, and yellow, respectively.}
  \label{tab:maxcity}
  \centering
  \tabcolsep=0.1cm
  \resizebox{0.72\textwidth}{!}{%
  \begin{tabular}{@{}l|lll|lll|lll@{}}
    \hline
     & \multicolumn{3}{c|}{MatrixCity} & \multicolumn{3}{c|}{Rubble} & \multicolumn{3}{c}{Building} \\
    \hline
    Metrics & SSIM$\uparrow$ & PSNR$\uparrow$ & LPIPS$\downarrow$ & SSIM$\uparrow$ & PSNR$\uparrow$ & LPIPS$\downarrow$ & SSIM$\uparrow$ & PSNR$\uparrow$ & LPIPS$\downarrow$ \\
    \hline
    Mega-NeRF~\pub{CVPR22}~\cite{exp_meganerf} & \makecell{\centering -} & \makecell{\centering -} & \makecell{\centering -} & 0.553 & 24.06 & 0.516 & 0.547 & 20.93 & 0.504 \\
    Switch-NeRF~\pub{ICLR22}~\cite{exp_switchnerf} & \makecell{\centering -} & \makecell{\centering -} & \makecell{\centering -} & 0.562 & 24.31 & 0.496 & 0.579 & 21.54 & 0.474 \\
    3DGS~\pub{ToG23}~\cite{relate_3dgs} & 0.737 & 23.79 & 0.384 & 0.746 & 24.30 & 0.324 & 0.676 & 20.11 & 0.372 \\
    GaMeS~\pub{arXiv24}~\cite{exp_games} & 0.773 & 24.64 & 0.367 & 0.767 & 24.64 & 0.311 & 0.698 & 20.97 & 0.329 \\
    Compact3DGS~\pub{CVPR24}~\cite{exp_compact3dgs} & 0.795 & 26.03 & 0.370 & 0.719 & 23.01 & 0.373 & 0.645 & 20.00 & 0.425 \\
    Scaffold-GS~\pub{CVPR24}~\cite{relate_scaffoldgs} & \cellcolor{s}0.842 & \cellcolor{s}27.25 & \cellcolor{t}0.290 & 0.736 & 23.71 & 0.346 & 0.711 & 21.55 & 0.328 \\
    Pixel-GS~\pub{ECCV24}~\cite{exp_pixelgs} & 0.814 & 25.60 & 0.324 & \cellcolor{t}0.771 & 24.67 & 0.308 & 0.700 & 21.00 & 0.352 \\
    MVGS~\pub{arxiv24}~\cite{mvgs} & \cellcolor{t}0.836 & \cellcolor{t}26.92 & \cellcolor{s}0.271 & \cellcolor{s}0.772 & \cellcolor{s}25.82 & \cellcolor{f}0.268 & \cellcolor{t}0.731 & \cellcolor{s}22.46 & \cellcolor{s}0.272 \\
    CityGS-X~\pub{ICCV25}~\cite{citygs-x} & 0.828 & 26.40 & 0.309 & \cellcolor{s}0.772 & \cellcolor{t}24.96 & \cellcolor{t}0.290 & \cellcolor{s}0.739 & \cellcolor{t}22.20 & \cellcolor{t}0.278 \\
    \hline
    Ours & \cellcolor{f}0.855 & \cellcolor{f}28.05 & \cellcolor{f}0.249 & \cellcolor{f}0.794 & \cellcolor{f}26.90 & \cellcolor{s}0.280 & \cellcolor{f}0.759 & \cellcolor{f}23.03 & \cellcolor{f}0.270 \\
    \hline
  \end{tabular}
  }
\end{table*}
\begin{figure*}[!t]
\centering
\includegraphics[width=1.0\textwidth]{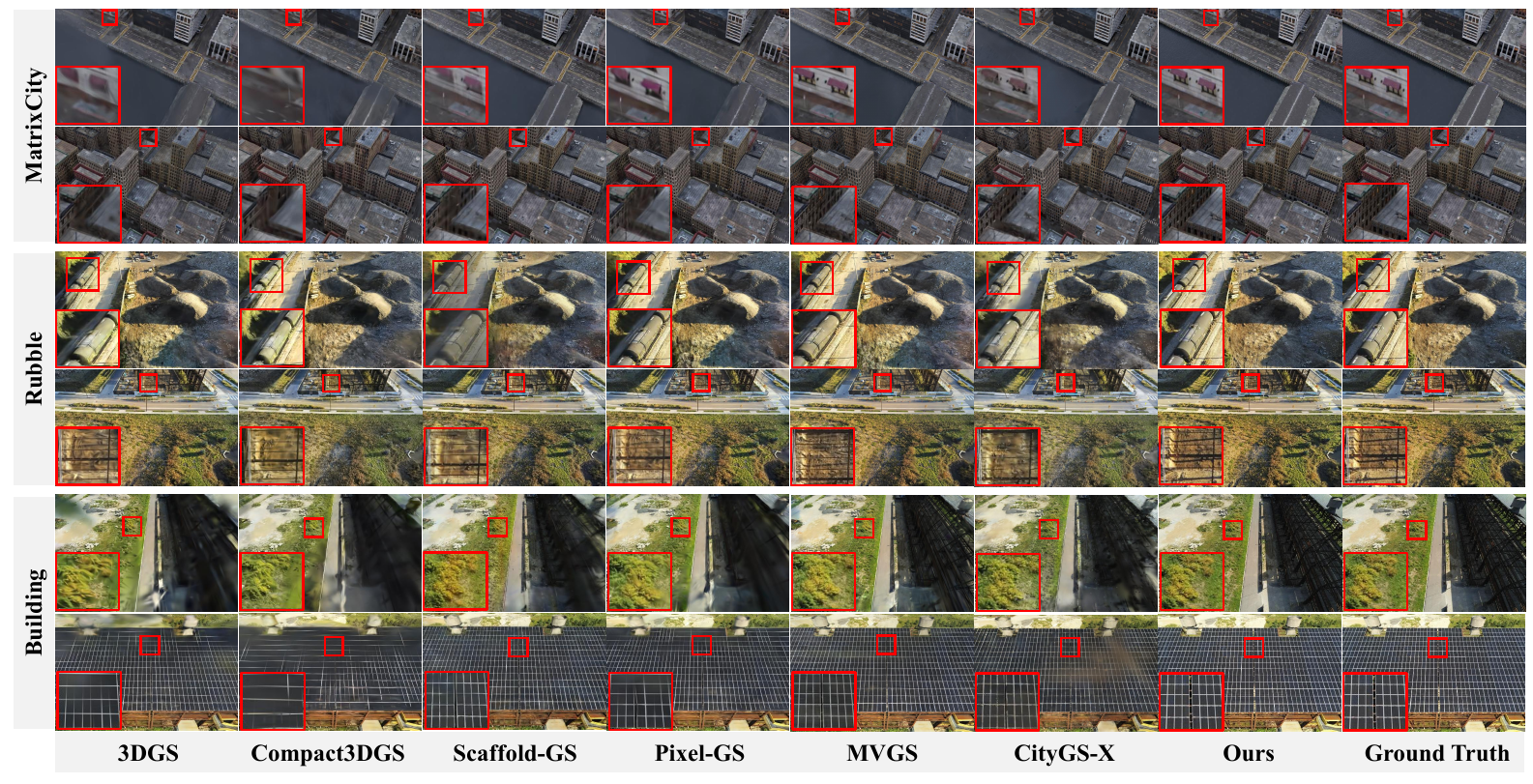}
\caption{Qualitative comparison of our method with existing advanced techniques on the MatrixCity and Mill-19 datasets. \textcolor{red}{Red} boxes are used to highlight visual differences. As illustrated, in the building scene, our method achieves high-fidelity fine-grained reconstruction of complex textures, such as the exterior of solar panels, surpassing other methods. Please \textbf{\emph{zoom in}} for more detailed visual results.}
\label{fig_mill}
\end{figure*}
\subsection{Datasets}
We assess the performance of our method by comparing it with state-of-the-art approaches on 13 outdoor scenes of diverse scales, which include eight scenes from four publicly available datasets and five scenes collected independently.

\textbf{Public Scenes.} We perform evaluations on eight scenes, which include the Rubble and Building from the Mill-19 dataset~\cite{exp_meganerf}, MatrixCity from the city-scale MatrixCity-Aerial dataset~\cite{exp_matrixcity}, and Train and Truck from the Tanks \& Temples dataset~\cite{exp_tnt}. Additionally, we select three particularly challenging aerial scenes from areas 1, 4, and 5 of the WHU dataset~\cite{exp_whu} to assess the performance of various methods.

\textbf{Self-collected Scenes.} To further evaluate the performance of our method in campus environments and plateau regions with large coverage areas and rich details, we collect the SCUT Campus Aerial (SCUT-CA) scenes, which consist of three scenes featuring teaching, office, and history museum buildings. Specifically, we use the DJI Matrice 300 RTK to capture images from approximately 100 meters above three distinct scenes at the Wushan Campus of South China University of Technology, with 125, 197, and 273 images, respectively, totaling 595 images. Each image has a resolution of 4082$\times$2148. To maintain consistency with the 3DGS~\cite{relate_3dgs} input, we downsample the images to a resolution of 1600$\times$841.

We also collect scenes from plateau regions, which include two scenes characterized by vast distribution, along with diverse geographical features. Specifically, we use the DJI Matrice 300 to capture images at an altitude of approximately 100 meters in the plateau region (Linzhi, Tibet), obtaining 203 and 170 images, respectively, for a total of 373 images. Each image has a resolution of 6222$\times$4148, and we downsample these images to a resolution of 1600$\times$841 as well. Camera parameters are estimated using widely adopted techniques such as COLMAP~\cite{exp_colmap}.

\subsection{Evaluation Metrics}
Following previous methods~\cite{relate_3dgs,exp_meganerf}, we primarily evaluate the quality of the rendered images using three metrics: PSNR, SSIM~\cite{exp_sim}, and the VGG implementation of LPIPS~\cite{exp_llism}.

\subsection{Implementation Details}
Our experiments are conducted on an NVIDIA 3090 GPU using PyTorch, with each scene trained for 30,000 iterations. In the CSCM, we progressively activate the process of acquiring HDE features at three granularities—coarse to fine—during training iterations 1, 12,000, and 21,000, respectively. At each level, the pixel resolution of the tri-plane feature maps is doubled compared to the previous level, while the density of the context grid is also increased by a factor of two. To optimize both computational efficiency and rendering performance, we apply the tri-plane spatial attention mechanism for enhancement solely at the first level's tri-plane feature maps. In our experiments, the MLPs used to update Gaussian attributes, as well as those in the CSCM, both consist of two layers, with the latter incorporating an additional batch normalization operation. 
All parameters related to Gaussian optimization and the densification process are set according to the original configurations from previous methods \cite{relate_scaffoldgs}.

\begin{figure*}[!t]
\centering
\includegraphics[width=1.0\textwidth]{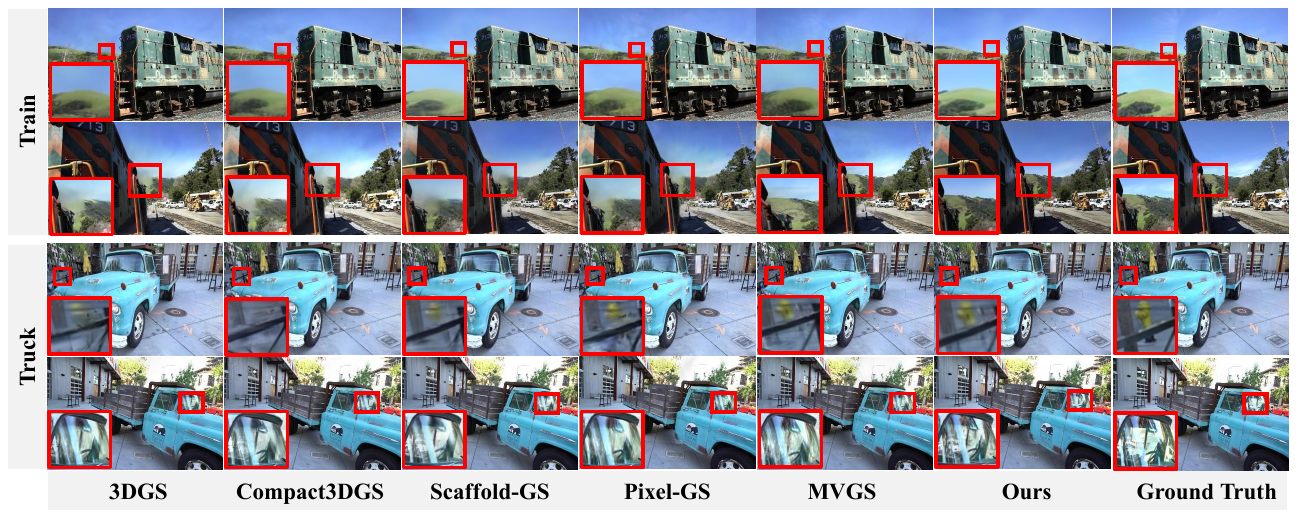}
\caption{Novel view synthesis results on the Tanks \& Temples dataset. Our method, leveraging the SVC module, achieves superior detail in challenging scenarios. This is evidenced by the sharper reconstruction of distant mountain edges in the train scene and the comprehensive modeling of glass reflection details in the truck scene, highlighting the efficacy of our approach.}
\label{fig_tnt}
\end{figure*}

\subsection{Result Analysis}
\subsubsection{Result on Mill-19 and MatrixCity dataset}
We rigorously evaluate our method through comprehensive quantitative and qualitative analyses. As shown in Table. \ref{tab:maxcity}, our approach achieves state-of-the-art performance across three scenarios: MatrixCity, Rubble and Building. In the Building scene, our approach achieves substantial gains over Scaffold-GS, with margins of +0.048 (SSIM), +1.48 (PSNR), and +0.058 (LPIPS), underscoring its superior rendering quality. We attribute this superiority to our collaborative framework's enhanced adaptability to spatially structured scenes - compared to Scaffold-GS's hash grid encoding, our architecture models geometric priors more effectively. Notably, Pixel-GS exhibits marked performance degradation in this scenario, suggesting that pixel-optimized point cloud processing struggles with large-scale environments containing complex spatial structures. For a fair comparison, we utilize the official open-source code of MVGS, which is based on the 3DGS framework. The sampled camera-image pair parameter $M$ in MVGS is also set to 4, maintaining consistency with our method. Our method demonstrably outperforms MVGS across the MatrixCity, Rubble, and Building scenes in most metrics. In the Rubble scene only, our method achieves an LPIPS score slightly below 0.012. The superior performance is primarily attributed to our SVC module, which mitigates the detail learning degradation issue in large-scale scenes, effectively preserving the ability to learn both the overall layout and local details. For a fair comparison, we also use the open-source code of CityGS-X and train it for the same number of iterations. Our method demonstrates superior performance to CityGS-X across all metrics. Fig. \ref{fig_mill} presents visual comparisons between our method and existing 3DGS baselines. The experimental results show that our approach consistently preserves high-fidelity detail reconstruction while effectively mitigating artifacts common in other methods. Particularly in the Building scene, our method achieves more accurate reconstruction of fine structural details such as solar panel grid patterns, demonstrating clear advantages over competing approaches.

\begin{table}[!h]
\caption{Quantitative comparison on Tanks \& Temples dataset. The best, second-best, and third-best results are marked in red, orange, and yellow, respectively.}
  \label{tab:tnt}
  \centering
  \tabcolsep=0.04cm
  \footnotesize  
  \setlength{\tabcolsep}{0.04cm}  
  \begin{tabular}{@{}l|lll|lll@{}}
    \hline
     & \multicolumn{3}{c|}{Train} & \multicolumn{3}{c}{Truck} \\
    \hline
    Metrics & SSIM & PSNR & LPIPS & SSIM & PSNR & LPIPS \\
    \hline
    Mip-NerF360~\cite{relate_mipnerf360} & 0.660 & 19.52 & 0.354 & 0.857 & 24.91 & 0.159 \\
    iNPG~\cite{exp_ingp} & 0.666 & 20.17 & 0.386 & 0.779 & 23.26 & 0.274 \\
    Plenoxels~\cite{exp_plenoxels} & 0.663 & 18.93 & 0.422 & 0.774 & 23.22 & 0.335 \\
    3DGS~\cite{relate_3dgs} & 0.815 & 22.07 & 0.208 & \cellcolor{s}0.882 & \cellcolor{t}25.46 & 0.147 \\
    GaMeS~\cite{exp_games} & \cellcolor{t}0.816 & 22.11 & 0.205 & 0.877 & 25.24 & 0.151 \\
    Compact3DGS~\cite{exp_compact3dgs} & 0.792 & 21.56 & 0.240 & 0.871 & 25.07 & 0.163 \\
    Scaffold-GS~\cite{relate_scaffoldgs} & \cellcolor{t}0.816 & \cellcolor{t}22.32 & 0.213 & \cellcolor{f}0.886 & \cellcolor{s}25.83 & \cellcolor{t}0.141 \\
    Pixel-GS~\cite{exp_pixelgs} & 0.813 & 22.01 & \cellcolor{t}0.188 & 0.872 & 25.03 & \cellcolor{t}0.131 \\
    MVGS~\cite{mvgs} & \cellcolor{s}0.833 & \cellcolor{s}22.43 & \cellcolor{f}0.155 & \cellcolor{t}0.881 & 25.45 & \cellcolor{f}0.116 \\
    \hline
    Ours & \cellcolor{f}0.840 & \cellcolor{f}23.76 & \cellcolor{s}0.158 & \cellcolor{t}0.881 & \cellcolor{f}26.09 &\cellcolor{s}0.126 \\
    \hline
  \end{tabular}
\end{table}
\begin{table*}[!t]
\caption{Quantitative Comparison on SCUT\_CA outdoor scenes. The best, second-best, and third-best results are marked in red, orange, and yellow, respectively.}
  \label{tab:scut_ca}
  \centering
  \tabcolsep=0.1cm
  \resizebox{0.72\textwidth}{!}{%
  \begin{tabular}{@{}l|lll|lll|lll@{}}
    \hline
     & \multicolumn{3}{c|}{Tower Two} & \multicolumn{3}{c|}{School History Museum} & \multicolumn{3}{c}{EI School} \\
    \hline
    Metrics & SSIM$\uparrow$ & PSNR$\uparrow$ & LPIPS$\downarrow$ & SSIM$\uparrow$ & PSNR$\uparrow$ & LPIPS$\downarrow$ & SSIM$\uparrow$ & PSNR$\uparrow$ & LPIPS$\downarrow$ \\
    \hline
    3DGS~\pub{ToG23}~\cite{relate_3dgs} & 0.887 & 29.87 & 0.143 & 0.849 & 28.54 & 0.189 & 0.882 & 29.12 & 0.132 \\
    GaMeS~\pub{arXiv24}~\cite{exp_games} & \cellcolor{t}0.891 & \cellcolor{t}29.99 & \cellcolor{t}0.140 & \cellcolor{t}0.861 & \cellcolor{t}28.83 & \cellcolor{t}0.182 & \cellcolor{t}0.889 & \cellcolor{t}29.36 & \cellcolor{t}0.120 \\
    Compact3DGS~\pub{CVPR24}~\cite{exp_compact3dgs} & 0.873 & 29.03 & 0.164 & 0.844 & 27.96 & 0.211 & 0.863 & 27.97 & 0.160 \\
    Scaffold-GS~\pub{CVPR24}~\cite{relate_scaffoldgs} & 0.880 & 29.49 & 0.155 & 0.854 & 28.44 & 0.197 & 0.873 & 28.67 & 0.147 \\
    Pixel-GS~\pub{ECCV24}~\cite{exp_pixelgs} & \cellcolor{s}0.915 & \cellcolor{s}30.81 & \cellcolor{f}0.097 & \cellcolor{s}0.882 & \cellcolor{s}29.34 & \cellcolor{s}0.142 & \cellcolor{s}0.909 & \cellcolor{s}29.77 & \cellcolor{s}0.092 \\
    \hline
    Ours & \cellcolor{f}0.916 & \cellcolor{f}31.25 & \cellcolor{s}0.100 & \cellcolor{f}0.887 & \cellcolor{f}29.87 & \cellcolor{f}0.130 & \cellcolor{f}0.912 & \cellcolor{f}30.53 & \cellcolor{f}0.090 \\
    \hline
  \end{tabular}
  }
\end{table*}
\begin{figure*}[!h]
\centering
\includegraphics[width=1.0\textwidth]{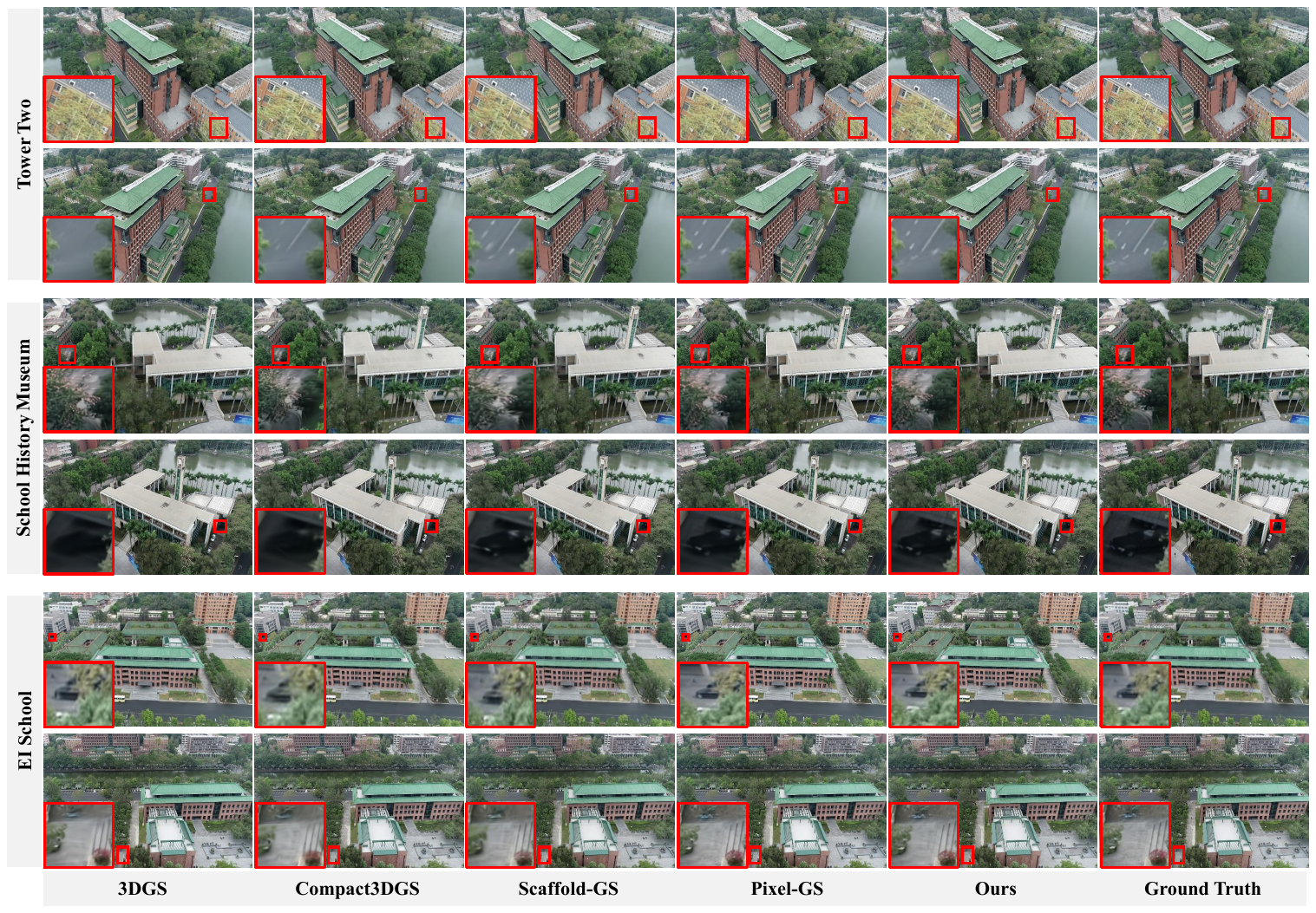}
\caption{Visualization results on our custom SCUT\_CA dataset. Highlighted regions (e.g., \textcolor{red}{red} boxes) reveal that our method renders significantly finer details and more realistic textures for elements such as leaves, lane markings, and distant objects, underscoring its superiority over competing approaches.}
\label{fig_scut_ca}
\end{figure*}
\subsubsection{Result on Tanks \& Temples dataset}
We present the quantitative results on Tanks and Temples in Table. \ref{tab:tnt}. Our method demonstrates superiority over other competitive approaches across most metrics on the 'Train' scene. Specifically, compared to Scaffold-GS, our approach achieves significant improvements of 0.024 in SSIM, 1.44 in PSNR, and 0.055 in LPIPS. For a fair comparison, the camera-image pair sampling parameter $M$ in MVGS is also set to 4. Our method achieves performance gains over MVGS, yielding improvements of $0.007$ and $1.33$ on the SSIM and PSNR metrics, respectively. We also note that our method falls marginally behind MVGS on the LPIPS metric for the 'Train' and 'Truck' scenes. As visually confirmed in Fig. \ref{fig_tnt}, our method effectively separates the background sky from distant mountains in the rendered imagery. For the 'Truck' scene, our method exhibits superior performance in both PSNR and LPIPS metrics compared to other approaches. While it shows a marginal 0.005 deficit in SSIM relative to Scaffold-GS, our method produces visually superior results. Specifically, it better reconstructs fire hydrant contours and significantly reduces artifacts in the truck's window regions.

\subsubsection{Result on SCUT-CA scenes}
We further evaluate our method's performance on our custom SCUT-CA dataset. As shown in Table. \ref{tab:scut_ca}, our approach demonstrates superior performance across all evaluation metrics in both the School History Museum and EI School scenes when compared to other baseline methods. While exhibiting a marginal 0.003 deficit in LPIPS when compared to Pixel-GS, our method shows significant advantages in both SSIM and PSNR metrics. Notably, Scaffold-GS exhibits substantial performance degradation on our custom dataset in comparison to public scene datasets, whereas our method maintains consistently robust performance. This enhanced robustness can be attributed to our SVC module. The comparative results in Fig. \ref{fig_scut_ca} reveal that the original 3DGS fails to reconstruct the zebra crossing details in the scene, while our method not only accurately recovers these fine details but also significantly reduces rendering artifacts. Furthermore, we observe that in the School History Museum scene, both 3DGS and Compact3DGS fail to effectively render vehicle wheels.

\begin{table*}[!t]
\caption{Quantitative Comparison on WHU dataset and plateau region scenes. The best, second-best, and third-best results are marked in red, orange, and yellow, respectively.}
  \label{tab:whu}
  \centering
  \tabcolsep=0.1cm
  \resizebox{1.0\textwidth}{!}{%
  \begin{tabular}{@{}l|lll|lll|lll|lll|lll@{}}
    \hline
     & \multicolumn{9}{c|}{WHU} & \multicolumn{6}{c}{Plateau Region} \\
    \hline
     & \multicolumn{3}{c|}{Area1} & \multicolumn{3}{c|}{Area4} & \multicolumn{3}{c|}{Area5} & \multicolumn{3}{c|}{Scene1}& \multicolumn{3}{c}{Scene2} \\
    \hline
    Metrics & SSIM$\uparrow$ & PSNR$\uparrow$ & LPIPS$\downarrow$ & SSIM$\uparrow$ & PSNR$\uparrow$ & LPIPS$\downarrow$ & SSIM$\uparrow$ & PSNR$\uparrow$ & LPIPS$\downarrow$& SSIM$\uparrow$ & PSNR$\uparrow$ & LPIPS$\downarrow$& SSIM$\uparrow$ & PSNR$\uparrow$ & LPIPS$\downarrow$ \\
    \hline
    3DGS~\pub{ToG23}~\cite{relate_3dgs} & 0.887 & 30.84 & 0.216 & 0.901 & 28.87 & 0.200 & 0.799 & 26.04 & 0.329 & 0.723 & 28.46 & \cellcolor{t}0.334 & \cellcolor{t}0.745 & 29.62 & \cellcolor{t}0.369 \\
    GaMeS~\pub{arXiv24}~\cite{exp_games} & 0.902 & 31.28 & \cellcolor{t}0.207 & 0.892 & 28.34 & 0.203 & 0.812 & 26.48 & 0.311 & \cellcolor{s}0.783 & \cellcolor{s}29.24 & \cellcolor{f}0.308 & \cellcolor{s}0.759 & 29.79 & \cellcolor{s}0.363 \\
    Compact3DGS~\pub{CVPR24}~\cite{exp_compact3dgs} & \cellcolor{t}0.922 & \cellcolor{t}32.69 & 0.217 & \cellcolor{t}0.946 & \cellcolor{t}33.79 & \cellcolor{t}0.147 & \cellcolor{t}0.865 & \cellcolor{t}28.37 & \cellcolor{t}0.273 & 0.721 & 28.32 & 0.355 & 0.721 & 29.78 & 0.400 \\
    Scaffold-GS~\pub{CVPR24}~\cite{relate_scaffoldgs} & \cellcolor{s}0.978 & \cellcolor{s}38.90 & \cellcolor{s}0.068 & \cellcolor{s}0.980 & \cellcolor{s}38.86 & \cellcolor{s}0.063 & \cellcolor{s}0.964 & \cellcolor{s}34.70 & \cellcolor{s}0.089 & \cellcolor{f}0.787 & \cellcolor{f}29.66 & \cellcolor{f}0.308 & 0.737 & \cellcolor{t}30.06 & 0.393 \\
    Pixel-GS~\pub{ECCV24}~\cite{exp_pixelgs} & 0.814 & 23.51 & 0.343 & 0.806 & 21.07 & 0.321 & 0.631 & 17.61 & 0.474 & 0.663 & 28.20 & 0.486 & 0.743 & \cellcolor{s}30.42 & 0.440 \\
    \hline
    Ours & \cellcolor{f}0.988 & \cellcolor{f}42.27 & \cellcolor{f}0.034 & \cellcolor{f}0.987 & \cellcolor{f}41.20 & \cellcolor{f}0.040 & \cellcolor{f}0.975 & \cellcolor{f}37.13 & \cellcolor{f}0.058 & \cellcolor{t}0.734 & \cellcolor{t}29.03 & \cellcolor{s}0.332 & \cellcolor{f}0.769 & \cellcolor{f}30.68 & \cellcolor{f}0.359 \\
    \hline
  \end{tabular}
  }
\end{table*}

\begin{figure*}[!t]
\centering
\includegraphics[width=1.0\textwidth]{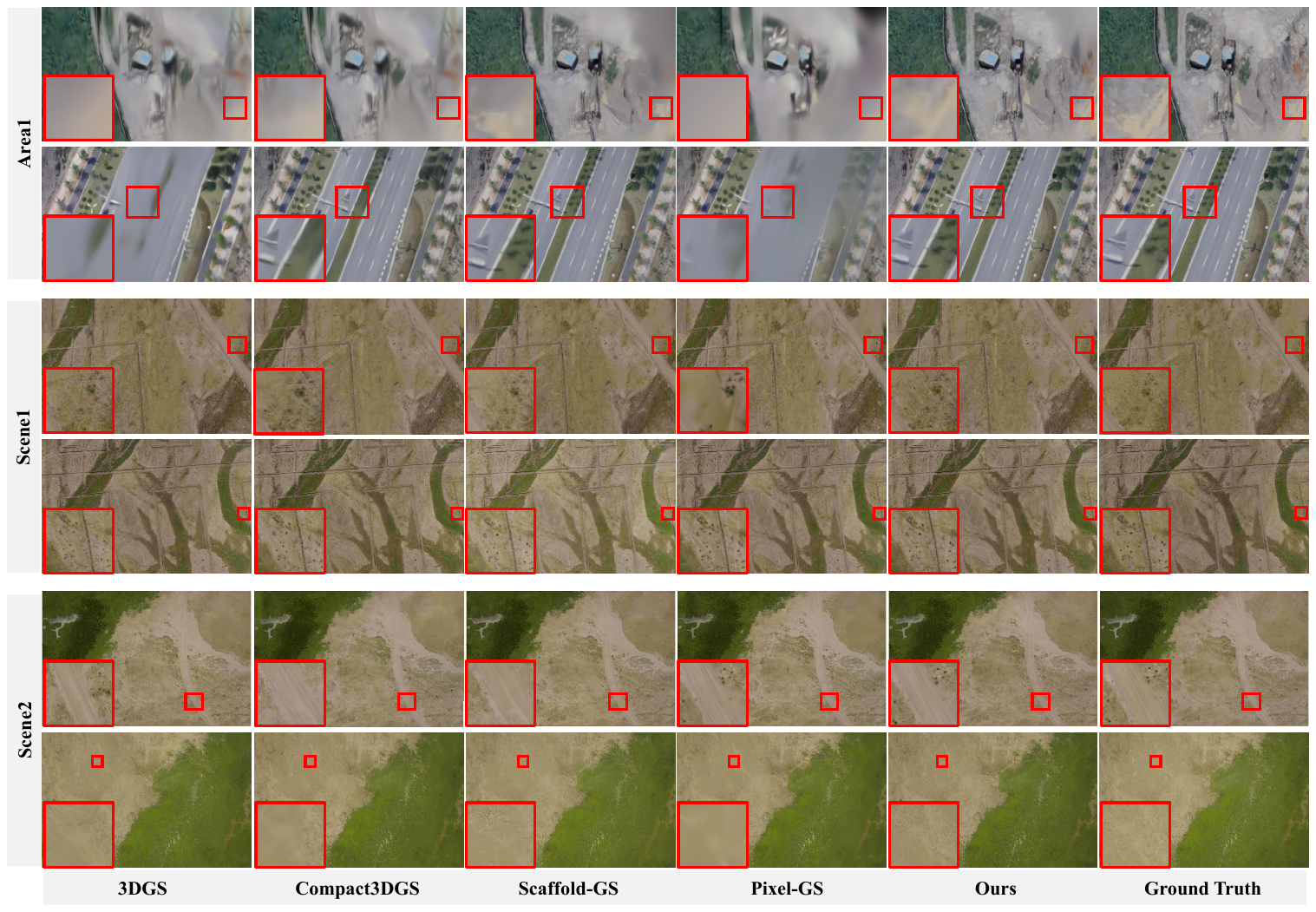}
\caption{Results on the WHU dataset and plateau region scenes, our method achieves highly realistic modeling of broad scene structures and precise capture of fine surface details. For instance, 'area1' showcases more authentic land and road structures. Simultaneously, our method accurately delineates subtle features like the distribution of vegetation on plateaus and the complex surface textures of sandy terrains, demonstrating superior structural awareness and detail retrieval.}
\label{fig_whu}
\end{figure*}

\subsubsection{Result on WHU dataset and Plateau Region scenes}
We systematically evaluate our method on both the WHU dataset and plateau scenes, with quantitative and qualitative results presented in Table. \ref{tab:whu} and Fig. \ref{fig_whu}, respectively. The WHU dataset is characterized by significant image blurring, while the plateau scenes exhibit distinct planar features. As demonstrated in Table. \ref{tab:whu}, our method maintains superior performance metrics even under challenging low-quality blurry conditions. Notably, in the plateau scene evaluation, our method achieves optimal results in Scene 2 but exhibits marginally degraded performance in Scene 1. Through rigorous analysis, we attribute the performance gap to non-uniform sampling during data acquisition in Scene 1, where the drone captured images at varying altitudes during its ascent. This inconsistency in viewpoint distribution disrupts the structure-view collaborative learning mechanism that our proposed SVC module relies on, ultimately leading to degraded model performance. Furthermore, our experiments reveal that mesh-based methods (e.g. GaMeS) demonstrate particularly strong performance in planar regions, owing to their inherent advantages in geometric representation using mesh structures. Visual comparisons in Fig. \ref{fig_whu} show that our method does not have artifacts, and contains more details with much higher accuracy than other methods.
\begin{table}[!t]
\centering
\caption{Ablation study on cross-structure collaborated module.\label{tab:ab_cs}}
\begin{tabular}{l|>{\columncolor{gray!20}}c>{\columncolor{gray!20}}c>{\columncolor{gray!20}}c}
\toprule
method & \cellcolor{gray!20}{SSIM $\uparrow$} & \cellcolor{gray!20}{PSNR $\uparrow$} & \cellcolor{gray!20}{LPIPS $\downarrow$} \\
\midrule
w/o tri-p. & 0.764 & 25.14 & 0.287\\
w/o cont. & 0.781 & 25.34 & 0.269 \\
w/o tri-p. atte. & 0.792 & 25.59 & 0.280 \\
full  & \textbf{0.803} & \textbf{25.99} & \textbf{0.266} \\
\bottomrule
\end{tabular}
\end{table}

\begin{figure*}[!t]
\centering
\includegraphics[width=1.0\textwidth]{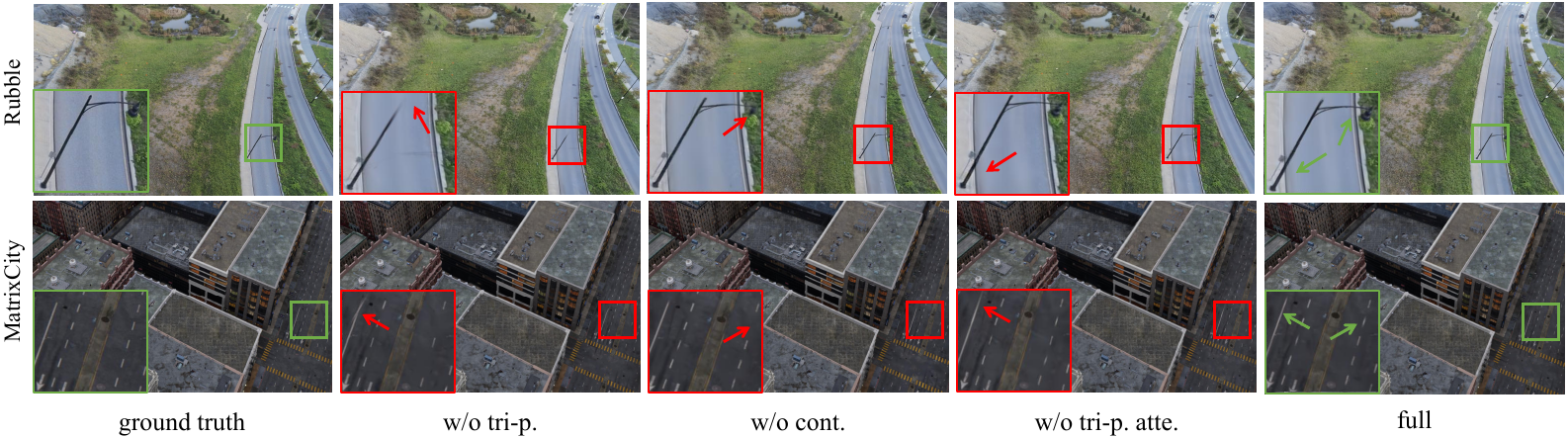}
\caption{Impact of the CSCM, demonstrated via ablation. \textcolor{red}{Red} and \textcolor{green}{green} boxes demarcate visual distinctions. By fusing attention-enhanced tri-plane features with context grid features, the module yields demonstrably more refined rendering and improved structural integrity. This stems from its enhanced ability to perceive both overarching point cloud architecture and fine-grained local details. Please \textbf{\emph{zoom in}} to view detailed results.}
\label{fig_ablation_s}
\end{figure*}
\subsection{Ablation Study}
\subsubsection{Effectiveness of Cross-Structure Collaboration Module}
We conduct an ablation study on our proposed CSCM under three configurations: (1) using only context grid encoding, (2) using only tri-plane feature encoding, and (3) combining both encodings. Additionally, we compare two feature fusion strategies: direct concatenation and attention-based fusion. As shown in Table. \ref{tab:ab_cs}, our experiments demonstrate that (1) fusing both encoding structures significantly improves performance, and (2) employing a lightweight attention mechanism further enhances the results. Qualitative results in Fig. \ref{fig_ablation_s} show that our CSCM generates rendered results with sharper line details.
\begin{table}[!h]
\centering
\caption{Ablation study on Cross-View Pruning Mechanism.\label{tab:ab_cvpm}}
\begin{tabular}{l|>{\columncolor{gray!20}}c>{\columncolor{gray!20}}c>{\columncolor{gray!20}}c}
\toprule
method & \cellcolor{gray!20}{SSIM $\uparrow$} & \cellcolor{gray!20}{PSNR $\uparrow$} & \cellcolor{gray!20}{LPIPS $\downarrow$} \\
\midrule
w/o \(L_\text{CVC}\) & 0.780 & 25.45 & 0.275 \\
w/o prune & 0.780 & 25.42 & 0.285 \\
full  & \textbf{0.803} & \textbf{25.99} & \textbf{0.266} \\
\bottomrule
\end{tabular}
\end{table}

\begin{figure}[!h]
\centering
\includegraphics[width=0.45\textwidth]{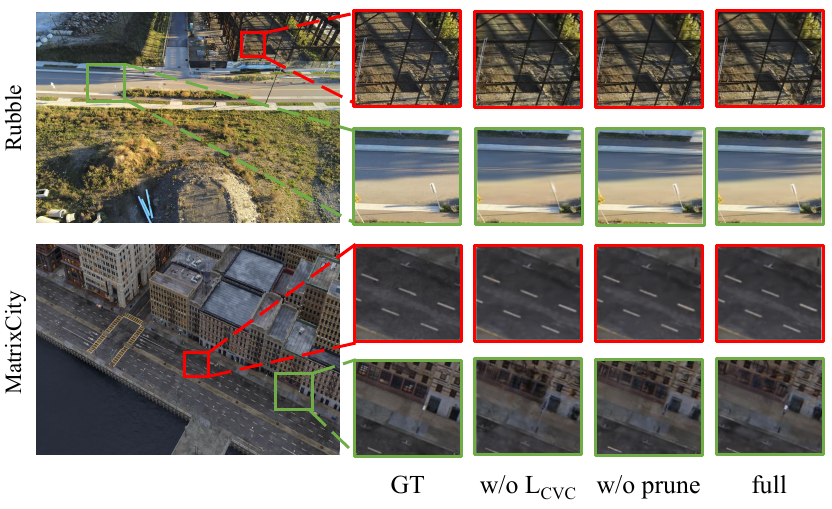}
\caption{Qualitative comparison of ablation study on CVPM. \textcolor{red}{Red} \& \textcolor{green}{green} squares highlight the visual differences for clearer comparison. By reconstructing the fine-grained structures in two scenes, such as the steel frame and signboards in the first row of images, and lane markings and street lamps in the second row of images, we demonstrate the superiority of the CVPM and validate the effectiveness of its individual components. Please \textbf{\emph{zoom in}} to view detailed results.}
\label{fig_ablation_v}
\end{figure}

\subsubsection{Effectiveness of Cross-View Pruning Mechanism}
In this section, we conduct ablation studies on the CVCL
and prune procedure in our CVPM. As shown in Table. \ref{tab:ab_cvpm}, integrating key components enables the model to achieve optimal performance, with particularly significant improvements in SSIM metrics. Qualitative results in Fig. \ref{fig_ablation_v} demonstrate that incorporating the components significantly enhances rendering quality by improving both detail clarity (e.g. street lamp) and artifact suppression.
	\section{Conclusion and future work}
	\label{sec:conclusion}
	In this paper, we introduce SplatCo, a structure-view collaborative framework specifically designed to address the scalability and fidelity challenges in rendering large-scale unbounded scenes. To overcome the inherent trade-off between global coherence and local details, we first propose the CSCM, which effectively harmonizes global tri-plane representations with local context grids via a novel hierarchical compensation strategy. Additionally, we implement the CVPM to mitigate Gaussian redundancy and visual artifacts, ensuring that only geometrically consistent primitives are retained. Finally, we incorporate the SVC module, which synergizes view gradients with structural gradients to robustly regularize Gaussian attributes. Extensive experiments on 13 diverse large-scale scenes demonstrate the superiority of our approach in rendering quality and detail preservation compared to state-of-the-art methods.

\textbf{Future Work}: In the future, we will explore reconstructing scenes with extreme structural characteristics, such as extensive textureless planes, by incorporating more robust geometric priors to enhance structural effectiveness, which is a worthwhile direction for achieving robust rendering of various challenging scenes.

\backmatter

\bibliography{SplatCo}

@article{citygs-x,
    title={CityGS-X: A Scalable Architecture for Efficient and Geometrically Accurate Large-Scale Scene Reconstruction}, 
    author={Yuanyuan Gao and Hao Li and Jiaqi Chen and Zhengyu Zou and Zhihang Zhong and Dingwen Zhang and Xiao Sun and Junwei Han},
    journal={arXiv preprint arXiv:2503.23044},
    year={2025},
    url={https://arxiv.org/abs/2503.23044}
}

@article{mvgs,
  title={Mvgs: Multi-view-regulated gaussian splatting for novel view synthesis},
  author={Du, Xiaobiao and Wang, Yida and Yu, Xin},
  journal={arXiv preprint arXiv:2410.02103},
  year={2024}
}

@ARTICLE{intro_vr,
  title={Efficient flower text entry in virtual reality},
  author={Leng, Jiaye and Wang, Lili and Liu, Xiaolong and Shi, Xuehuai and Wang, Miao},
  journal={IEEE Transactions on Visualization and Computer Graphics},
  volume={28},
  number={11},
  pages={3662--3672},
  year={2022},
  publisher={IEEE}
}

@inproceedings{intro_culture,
  title={Neural Radiance Fields (NeRF) for Multi-Scale 3D Modeling of Cultural Heritage Artifacts},
  author={Croce, V. and Malinverni, E. S. and Pierdicca, R. and Paolanti, M. and Frontoni, E. and Di Stefano, F. and Pansoni, S. and Tiribelli, S. and Malinverni, E. S. and Giovanola, B.},
  booktitle={29th CIPA Symposium},
  year={2024}
}

@Article{intro_twin,
  title={Embodied intelligence via learning and evolution},
  author={Gupta, Agrim and Savarese, Silvio and Ganguli, Surya and Fei-Fei, Li},
  journal={Nature communications},
  volume={12},
  number={1},
  pages={5721},
  year={2021},
  publisher={Nature Publishing Group UK London}
}

@Article{intro_driving,
  title={Visual evaluation for autonomous driving},
  author={Hou, Yijie and Wang, Chengshun and Wang, Junhong and Xue, Xiangyang and Zhang, Xiaolong Luke and Zhu, Jun and Wang, Dongliang and Chen, Siming},
  journal={IEEE Transactions on Visualization and Computer Graphics},
  volume={28},
  number={1},
  pages={1030--1039},
  year={2021},
  publisher={IEEE}
}

@inproceedings{relate_new_cui2017hsfm,
  title={HSfM: Hybrid structure-from-motion},
  author={Cui, Hainan and Gao, Xiang and Shen, Shuhan and Hu, Zhanyi},
  booktitle={Proceedings of the IEEE conference on computer vision and pattern recognition},
  pages={1212--1221},
  year={2017}
}

@inproceedings{relate_new_schonberger2016structure,
  title={Structure-from-motion revisited},
  author={Schonberger, Johannes L and Frahm, Jan-Michael},
  booktitle={Proceedings of the IEEE conference on computer vision and pattern recognition},
  pages={4104--4113},
  year={2016}
}

@article{relate_new_zheng2017sift,
  title={SIFT meets CNN: A decade survey of instance retrieval},
  author={Zheng, Liang and Yang, Yi and Tian, Qi},
  journal={IEEE transactions on pattern analysis and machine intelligence},
  volume={40},
  number={5},
  pages={1224--1244},
  year={2017},
  publisher={IEEE}
}

@inproceedings{relate_new_agarwal2010bundle,
  title={Bundle adjustment in the large},
  author={Agarwal, Sameer and Snavely, Noah and Seitz, Steven M and Szeliski, Richard},
  booktitle={Computer Vision--ECCV 2010: 11th European Conference on Computer Vision, Heraklion, Crete, Greece, September 5-11, 2010, Proceedings, Part II 11},
  pages={29--42},
  year={2010},
  organization={Springer}
}

@inproceedings{relate_new_mesh_liu2019liquid,
  title={Liquid warping gan: A unified framework for human motion imitation, appearance transfer and novel view synthesis},
  author={Liu, Wen and Piao, Zhixin and Min, Jie and Luo, Wenhan and Ma, Lin and Gao, Shenghua},
  booktitle={Proceedings of the IEEE/CVF international conference on computer vision},
  pages={5904--5913},
  year={2019}
}

@inproceedings{realet_new_mesh_guo2023vmesh,
  title={Vmesh: Hybrid volume-mesh representation for efficient view synthesis},
  author={Guo, Yuan-Chen and Cao, Yan-Pei and Wang, Chen and He, Yu and Shan, Ying and Zhang, Song-Hai},
  booktitle={SIGGRAPH Asia 2023 Conference Papers},
  pages={1--11},
  year={2023}
}

@article{relate_new_voxel_sun2023vgos,
  title={Vgos: Voxel grid optimization for view synthesis from sparse inputs},
  author={Sun, Jiakai and Zhang, Zhanjie and Chen, Jiafu and Li, Guangyuan and Ji, Boyan and Zhao, Lei and Xing, Wei and Lin, Huaizhong},
  journal={arXiv preprint arXiv:2304.13386},
  year={2023}
}

@article{relate_new_voxel_schwarz2022voxgraf,
  title={Voxgraf: Fast 3d-aware image synthesis with sparse voxel grids},
  author={Schwarz, Katja and Sauer, Axel and Niemeyer, Michael and Liao, Yiyi and Geiger, Andreas},
  journal={Advances in Neural Information Processing Systems},
  volume={35},
  pages={33999--34011},
  year={2022}
}

@article{relate_new_pointcloud_xiao2023distinguishing,
  title={Distinguishing and matching-aware unsupervised point cloud completion},
  author={Xiao, Haihong and Li, Yuqiong and Kang, Wenxiong and Wu, Qiuxia},
  journal={IEEE Transactions on Circuits and Systems for Video Technology},
  volume={33},
  number={9},
  pages={5160--5173},
  year={2023},
  publisher={IEEE}
}

@inproceedings{relate_tra_1,
  title={Massively parallel multiview stereopsis by surface normal diffusion},
  author={Galliani, Silvano and Lasinger, Katrin and Schindler, Konrad},
  booktitle={Proceedings of the IEEE international conference on computer vision},
  pages={873--881},
  year={2015}
}

@article{relate_tra_2,
  title={Multi-view stereo reconstruction of dense shape and complex appearance},
  author={Jin, Hailin and Soatto, Stefano and Yezzi, Anthony J},
  journal={International Journal of Computer Vision},
  volume={63},
  pages={175--189},
  year={2005},
  publisher={Springer}
}

@inproceedings{relate_mvsnet,
  title={Mvsnet: Depth inference for unstructured multi-view stereo},
  author={Yao, Yao and Luo, Zixin and Li, Shiwei and Fang, Tian and Quan, Long},
  booktitle={European conference on computer vision},
  pages={767--783},
  year={2018}
}

@inproceedings{relate_mvster,
  title={MVSTER: Epipolar transformer for efficient multi-view stereo},
  author={Wang, Xiaofeng and Zhu, Zheng and Huang, Guan and Qin, Fangbo and Ye, Yun and He, Yijia and Chi, Xu and Wang, Xingang},
  booktitle={European conference on computer vision},
  pages={573--591},
  year={2022},
  organization={Springer}
}

@article{relate_unimvsnet,
  title={Ea-mvsnet: learning error-awareness for enhanced multi-view stereo},
  author={Gu, Wencong and Xiao, Haihong and Zhao, Xueyan and Kang, Wenxiong},
  journal={IEEE Transactions on Circuits and Systems for Video Technology},
  year={2024},
  publisher={IEEE}
}

@inproceedings{relate_transmvsnet,
  title={Transmvsnet: Global context-aware multi-view stereo network with transformers},
  author={Ding, Yikang and Yuan, Wentao and Zhu, Qingtian and Zhang, Haotian and Liu, Xiangyue and Wang, Yuanjiang and Liu, Xiao},
  booktitle={Proceedings of the IEEE/CVF conference on computer vision and pattern recognition},
  pages={8585--8594},
  year={2022}
}

@inproceedings{relate_rmvsnet,
  title={Recurrent mvsnet for high-resolution multi-view stereo depth inference},
  author={Yao, Yao and Luo, Zixin and Li, Shiwei and Shen, Tianwei and Fang, Tian and Quan, Long},
  booktitle={Proceedings of the IEEE/CVF conference on computer vision and pattern recognition},
  pages={5525--5534},
  year={2019}
}

@inproceedings{relate_casmvsnet,
  title={Cascade cost volume for high-resolution multi-view stereo and stereo matching},
  author={Gu, Xiaodong and Fan, Zhiwen and Zhu, Siyu and Dai, Zuozhuo and Tan, Feitong and Tan, Ping},
  booktitle={Proceedings of the IEEE/CVF conference on computer vision and pattern recognition},
  pages={2495--2504},
  year={2020}
}

@inproceedings{relate_nerf,
  title={Nerf in the wild: Neural radiance fields for unconstrained photo collections},
  author={Martin-Brualla, Ricardo and Radwan, Noha and Sajjadi, Mehdi SM and Barron, Jonathan T and Dosovitskiy, Alexey and Duckworth, Daniel},
  booktitle={Proceedings of the IEEE/CVF conference on computer vision and pattern recognition},
  pages={7210--7219},
  year={2021}
}

@inproceedings{relate_new_li2023nerfacc,
  title={Nerfacc: Efficient sampling accelerates nerfs},
  author={Li, Ruilong and Gao, Hang and Tancik, Matthew and Kanazawa, Angjoo},
  booktitle={Proceedings of the IEEE/CVF international conference on computer vision},
  pages={18537--18546},
  year={2023}
}

@inproceedings{relate_new_korhonen2024efficient,
  title={Efficient NeRF Optimization-Not All Samples Remain Equally Hard},
  author={Korhonen, Juuso and Rangu, Goutham and Tavakoli, Hamed R and Kannala, Juho},
  booktitle={European Conference on Computer Vision},
  pages={198--213},
  year={2024},
  organization={Springer}
}

@inproceedings{relate_mipnerf,
  title={Mip-nerf: A multiscale representation for anti-aliasing neural radiance fields},
  author={Barron, Jonathan T and Mildenhall, Ben and Tancik, Matthew and Hedman, Peter and Martin-Brualla, Ricardo and Srinivasan, Pratul P},
  booktitle={Proceedings of the IEEE/CVF international conference on computer vision},
  pages={5855--5864},
  year={2021}
}

@inproceedings{relate_new_hu2023multiscale,
  title={Multiscale representation for real-time anti-aliasing neural rendering},
  author={Hu, Dongting and Zhang, Zhenkai and Hou, Tingbo and Liu, Tongliang and Fu, Huan and Gong, Mingming},
  booktitle={Proceedings of the IEEE/CVF International Conference on Computer Vision},
  pages={17772--17783},
  year={2023}
}

@inproceedings{relate_dnerf,
  title={D-nerf: Neural radiance fields for dynamic scenes},
  author={Pumarola, Albert and Corona, Enric and Pons-Moll, Gerard and Moreno-Noguer, Francesc},
  booktitle={Proceedings of the IEEE/CVF Conference on Computer Vision and Pattern Recognition},
  pages={10318--10327},
  year={2021}
}

@inproceedings{relate_deblur_nerf,
  title={Deblur-nerf: Neural radiance fields from blurry images},
  author={Ma, Li and Li, Xiaoyu and Liao, Jing and Zhang, Qi and Wang, Xuan and Wang, Jue and Sander, Pedro V},
  booktitle={Proceedings of the IEEE/CVF Conference on Computer Vision and Pattern Recognition},
  pages={12861--12870},
  year={2022}
}

@article{relate_new_wang2024mp,
  title={MP-NeRF: More refined deblurred neural radiance field for 3D reconstruction of blurred images},
  author={Wang, Xiaohui and Yin, Zhenyu and Zhang, Feiqing and Feng, Dan and Wang, Zisong},
  journal={Knowledge-Based Systems},
  volume={290},
  pages={111571},
  year={2024},
  publisher={Elsevier}
}

@article{relate_3dgs,
  title={3D Gaussian Splatting for Real-Time Radiance Field Rendering.},
  author={Kerbl, Bernhard and Kopanas, Georgios and Leimk{\"u}hler, Thomas and Drettakis, George},
  journal={ACM Trans. Graph.},
  volume={42},
  number={4},
  pages={139--1},
  year={2023}
}

@article{relate_new_pryadilshchikov2024t,
  title={T-3DGS: Removing Transient Objects for 3D Scene Reconstruction},
  author={Pryadilshchikov, Vadim and Markin, Alexander and Komarichev, Artem and Rakhimov, Ruslan and Wonka, Peter and Burnaev, Evgeny},
  journal={arXiv preprint arXiv:2412.00155},
  year={2024}
}

@inproceedings{relate_new_sun2024f,
  title={F-3dgs: Factorized coordinates and representations for 3d gaussian splatting},
  author={Sun, Xiangyu and Lee, Joo Chan and Rho, Daniel and Ko, Jong Hwan and Ali, Usman and Park, Eunbyung},
  booktitle={Proceedings of the 32nd ACM International Conference on Multimedia},
  pages={7957--7965},
  year={2024}
}

@article{relate_new_sakamiya2024avatarperfect,
  title={AvatarPerfect: User-Assisted 3D Gaussian Splatting Avatar Refinement with Automatic Pose Suggestion},
  author={Sakamiya, Jotaro and Shen, I and Zhang, Jinsong and Dogan, Mustafa Doga and Igarashi, Takeo and others},
  journal={arXiv preprint arXiv:2412.15609},
  year={2024}
}

@inproceedings{relate_new_chen2025thgs,
  title={THGS: Lifelike T alking H uman Avatar Synthesis From Monocular Video Via 3D G aussian S platting},
  author={Chen, Chuang and Yu, Lingyun and Yang, Quanwei and Zheng, Aihua and Xie, Hongtao},
  booktitle={Computer Graphics Forum},
  pages={e15282},
  year={2025},
  organization={Wiley Online Library}
}

@article{relate_new_qiu2024anigs,
  title={AniGS: Animatable Gaussian Avatar from a Single Image with Inconsistent Gaussian Reconstruction},
  author={Qiu, Lingteng and Zhu, Shenhao and Zuo, Qi and Gu, Xiaodong and Dong, Yuan and Zhang, Junfei and Xu, Chao and Li, Zhe and Yuan, Weihao and Bo, Liefeng and others},
  journal={arXiv preprint arXiv:2412.02684},
  year={2024}
}

@inproceedings{relate_new_lu2024scaffold,
  title={Scaffold-gs: Structured 3d gaussians for view-adaptive rendering},
  author={Lu, Tao and Yu, Mulin and Xu, Linning and Xiangli, Yuanbo and Wang, Limin and Lin, Dahua and Dai, Bo},
  booktitle={Proceedings of the IEEE/CVF Conference on Computer Vision and Pattern Recognition},
  pages={20654--20664},
  year={2024}
}

@article{chen2024dogs,
  title={Dogs: Distributed-oriented gaussian splatting for large-scale 3d reconstruction via gaussian consensus},
  author={Chen, Yu and Lee, Gim Hee},
  journal={Advances in Neural Information Processing Systems},
  volume={37},
  pages={34487--34512},
  year={2024}
}

@inproceedings{lin2024vastgaussian,
  title={Vastgaussian: Vast 3d gaussians for large scene reconstruction},
  author={Lin, Jiaqi and Li, Zhihao and Tang, Xiao and Liu, Jianzhuang and Liu, Shiyong and Liu, Jiayue and Lu, Yangdi and Wu, Xiaofei and Xu, Songcen and Yan, Youliang and others},
  booktitle={Proceedings of the IEEE/CVF Conference on Computer Vision and Pattern Recognition},
  pages={5166--5175},
  year={2024}
}

@article{kerbl2024hierarchical,
  title={A hierarchical 3d gaussian representation for real-time rendering of very large datasets},
  author={Kerbl, Bernhard and Meuleman, Andreas and Kopanas, Georgios and Wimmer, Michael and Lanvin, Alexandre and Drettakis, George},
  journal={ACM Transactions on Graphics (TOG)},
  volume={43},
  number={4},
  pages={1--15},
  year={2024},
  publisher={ACM New York, NY, USA}
}

@article{ren2024octree,
  title={Octree-gs: Towards consistent real-time rendering with lod-structured 3d gaussians},
  author={Ren, Kerui and Jiang, Lihan and Lu, Tao and Yu, Mulin and Xu, Linning and Ni, Zhangkai and Dai, Bo},
  journal={arXiv preprint arXiv:2403.17898},
  year={2024}
}

@inproceedings{liu2024citygaussian,
  title={Citygaussian: Real-time high-quality large-scale scene rendering with gaussians},
  author={Liu, Yang and Luo, Chuanchen and Fan, Lue and Wang, Naiyan and Peng, Junran and Zhang, Zhaoxiang},
  booktitle={European Conference on Computer Vision},
  pages={265--282},
  year={2024},
  organization={Springer}
}

@article{relate_gsdf,
  title={Gsdf: 3dgs meets sdf for improved rendering and reconstruction},
  author={Yu, Mulin and Lu, Tao and Xu, Linning and Jiang, Lihan and Xiangli, Yuanbo and Dai, Bo},
  journal={arXiv preprint arXiv:2403.16964},
  year={2024}
}

@inproceedings{relate_scaffoldgs,
  title={Scaffold-gs: Structured 3d gaussians for view-adaptive rendering},
  author={Lu, Tao and Yu, Mulin and Xu, Linning and Xiangli, Yuanbo and Wang, Limin and Lin, Dahua and Dai, Bo},
  booktitle={Proceedings of the IEEE/CVF Conference on Computer Vision and Pattern Recognition},
  pages={20654--20664},
  year={2024}
}

@inproceedings{relate_mipnerf360,
  title={Mip-nerf 360: Unbounded anti-aliased neural radiance fields},
  author={Barron, Jonathan T and Mildenhall, Ben and Verbin, Dor and Srinivasan, Pratul P and Hedman, Peter},
  booktitle={Proceedings of the IEEE/CVF conference on computer vision and pattern recognition},
  pages={5470--5479},
  year={2022}
}

@inproceedings{relate_dust3r,
  title={Dust3r: Geometric 3d vision made easy},
  author={Wang, Shuzhe and Leroy, Vincent and Cabon, Yohann and Chidlovskii, Boris and Revaud, Jerome},
  booktitle={Proceedings of the IEEE/CVF Conference on Computer Vision and Pattern Recognition},
  pages={20697--20709},
  year={2024}
}

@inproceedings{relate_must3r,
  title={Grounding image matching in 3d with mast3r},
  author={Leroy, Vincent and Cabon, Yohann and Revaud, J{\'e}r{\^o}me},
  booktitle={European Conference on Computer Vision},
  pages={71--91},
  year={2024},
  organization={Springer}
}

@article{relate_fast3r,
  title={Fast3R: Towards 3D Reconstruction of 1000+ Images in One Forward Pass},
  author={Yang, Jianing and Sax, Alexander and Liang, Kevin J and Henaff, Mikael and Tang, Hao and Cao, Ang and Chai, Joyce and Meier, Franziska and Feiszli, Matt},
  journal={arXiv preprint arXiv:2501.13928},
  year={2025}
}

@article{relate_d2ust3r,
  title={D\^{} 2USt3R: Enhancing 3D Reconstruction with 4D Pointmaps for Dynamic Scenes},
  author={Han, Jisang and An, Honggyu and Jung, Jaewoo and Narihira, Takuya and Seo, Junyoung and Fukuda, Kazumi and Kim, Chaehyun and Hong, Sunghwan and Mitsufuji, Yuki and Kim, Seungryong},
  journal={arXiv preprint arXiv:2504.06264},
  year={2025}
}

@inproceedings{relate_gsrec,
  title={Surface reconstruction from 3d gaussian splatting via local structural hints},
  author={Wu, Qianyi and Zheng, Jianmin and Cai, Jianfei},
  booktitle={European Conference on Computer Vision},
  pages={441--458},
  year={2024},
  organization={Springer}
}

@article{relate_gsroom,
      title={GaussianRoom: Improving 3D Gaussian Splatting with SDF Guidance and Monocular Cues for Indoor Scene Reconstruction}, 
      author={Haodong Xiang and Xinghui Li and Kai Cheng and Xiansong Lai and Wanting Zhang and Zhichao Liao and Long Zeng and Xueping Liu},
      year={2025},
      journal={arXiv preprint arXiv:2504.06264},
}

@inproceedings{method_hash_grid,
  title={Shacira: Scalable hash-grid compression for implicit neural representations},
  author={Girish, Sharath and Shrivastava, Abhinav and Gupta, Kamal},
  booktitle={Proceedings of the IEEE/CVF International Conference on Computer Vision},
  pages={17513--17524},
  year={2023}
}

@article{method_fpn,
  title={Multi-dimensional graph interactional network for progressive point cloud completion},
  author={Xiao, Haihong and Xu, Hongbin and Li, Yuqiong and Kang, Wenxiong},
  journal={IEEE Transactions on Instrumentation and Measurement},
  volume={72},
  pages={1--12},
  year={2022},
  publisher={IEEE}
}

@inproceedings{method_kplanes,
  title={K-planes: Explicit radiance fields in space, time, and appearance},
  author={Fridovich-Keil, Sara and Meanti, Giacomo and Warburg, Frederik Rahb{\ae}k and Recht, Benjamin and Kanazawa, Angjoo},
  booktitle={Proceedings of the IEEE/CVF Conference on Computer Vision and Pattern Recognition},
  pages={12479--12488},
  year={2023}
}

@inproceedings{method_cbam,
  title={Cbam: Convolutional block attention module},
  author={Woo, Sanghyun and Park, Jongchan and Lee, Joon-Young and Kweon, In So},
  booktitle={Proceedings of the European conference on computer vision (ECCV)},
  pages={3--19},
  year={2018}
}

@article{method_contextgs,
  title={Contextgs: Compact 3d gaussian splatting with anchor level context model},
  author={Wang, Yufei and Li, Zhihao and Guo, Lanqing and Yang, Wenhan and Kot, Alex and Wen, Bihan},
  journal={Advances in neural information processing systems},
  volume={37},
  pages={51532--51551},
  year={2024}
}

@inproceedings{method_trigs,
  title={Triplane meets gaussian splatting: Fast and generalizable single-view 3d reconstruction with transformers},
  author={Zou, Zi-Xin and Yu, Zhipeng and Guo, Yuan-Chen and Li, Yangguang and Liang, Ding and Cao, Yan-Pei and Zhang, Song-Hai},
  booktitle={Proceedings of the IEEE/CVF conference on computer vision and pattern recognition},
  pages={10324--10335},
  year={2024}
}

@article{method_igs,
  title={Implicit gaussian splatting with efficient multi-level tri-plane representation},
  author={Wu, Minye and Tuytelaars, Tinne},
  journal={arXiv preprint arXiv:2408.10041},
  year={2024}
}

@inproceedings{exp_meganerf,
  title={Mega-nerf: Scalable construction of large-scale nerfs for virtual fly-throughs},
  author={Turki, Haithem and Ramanan, Deva and Satyanarayanan, Mahadev},
  booktitle={Proceedings of the IEEE/CVF Conference on Computer Vision and Pattern Recognition},
  pages={12922--12931},
  year={2022}
}

@inproceedings{exp_matrixcity,
  title={Matrixcity: A large-scale city dataset for city-scale neural rendering and beyond},
  author={Li, Yixuan and Jiang, Lihan and Xu, Linning and Xiangli, Yuanbo and Wang, Zhenzhi and Lin, Dahua and Dai, Bo},
  booktitle={Proceedings of the IEEE/CVF International Conference on Computer Vision},
  pages={3205--3215},
  year={2023}
}

@article{exp_tnt,
  title={Tanks and temples: Benchmarking large-scale scene reconstruction},
  author={Knapitsch, Arno and Park, Jaesik and Zhou, Qian-Yi and Koltun, Vladlen},
  journal={ACM Transactions on Graphics},
  volume={36},
  number={4},
  pages={1--13},
  year={2017},
  publisher={ACM New York, NY, USA}
}

@inproceedings{exp_whu,
  title={A novel recurrent encoder-decoder structure for large-scale multi-view stereo reconstruction from an open aerial dataset},
  author={Liu, Jin and Ji, Shunping},
  booktitle={Proceedings of the IEEE/CVF conference on computer vision and pattern recognition},
  pages={6050--6059},
  year={2020}
}

@article{exp_colmap,
  title={COLMAP-SLAM: A framework for visual odometry},
  author={Morelli, Luca and Ioli, Francesco and Beber, Raniero and Menna, Fabio and Remondino, Fabio and Vitti, Alfonso},
  journal={The International Archives of the Photogrammetry, Remote Sensing and Spatial Information Sciences},
  volume={48},
  pages={317--324},
  year={2023},
  publisher={Copernicus Publications G{\"o}ttingen, Germany}
}

@article{exp_sim,
  title={Image quality assessment: from error visibility to structural similarity},
  author={Wang, Zhou and Bovik, Alan C and Sheikh, Hamid R and Simoncelli, Eero P},
  journal={IEEE transactions on image processing},
  volume={13},
  number={4},
  pages={600--612},
  year={2004},
  publisher={IEEE}
}

@inproceedings{exp_llism,
  title={The unreasonable effectiveness of deep features as a perceptual metric},
  author={Zhang, Richard and Isola, Phillip and Efros, Alexei A and Shechtman, Eli and Wang, Oliver},
  booktitle={Proceedings of the IEEE conference on computer vision and pattern recognition},
  pages={586--595},
  year={2018}
}

@article{exp_ingp,
  title={Instant neural graphics primitives with a multiresolution hash encoding},
  author={M{\"u}ller, Thomas and Evans, Alex and Schied, Christoph and Keller, Alexander},
  journal={ACM transactions on graphics (TOG)},
  volume={41},
  number={4},
  pages={1--15},
  year={2022},
  publisher={ACM New York, NY, USA}
}

@inproceedings{exp_plenoxels,
  title={Plenoxels: Radiance fields without neural networks},
  author={Fridovich-Keil, Sara and Yu, Alex and Tancik, Matthew and Chen, Qinhong and Recht, Benjamin and Kanazawa, Angjoo},
  booktitle={Proceedings of the IEEE/CVF conference on computer vision and pattern recognition},
  pages={5501--5510},
  year={2022}
}

@article{exp_games,
  title={Games: Mesh-based adapting and modification of gaussian splatting},
  author={Waczy{\'n}ska, Joanna and Borycki, Piotr and Tadeja, S{\l}awomir and Tabor, Jacek and Spurek, Przemys{\l}aw},
  journal={arXiv preprint arXiv:2402.01459},
  year={2024}
}

@inproceedings{exp_compact3dgs,
  title={Compact 3d gaussian representation for radiance field},
  author={Lee, Joo Chan and Rho, Daniel and Sun, Xiangyu and Ko, Jong Hwan and Park, Eunbyung},
  booktitle={Proceedings of the IEEE/CVF Conference on Computer Vision and Pattern Recognition},
  pages={21719--21728},
  year={2024}
}

@inproceedings{exp_switchnerf,
  title={Switch-nerf: Learning scene decomposition with mixture of experts for large-scale neural radiance fields},
  author={Zhenxing, MI and Xu, Dan},
  booktitle={The Eleventh International Conference on Learning Representations},
  year={2022}
}

@article{exp_pixelgs,
  title={Pixel-gs: Density control with pixel-aware gradient for 3d gaussian splatting},
  author={Zhang, Zheng and Hu, Wenbo and Lao, Yixing and He, Tong and Zhao, Hengshuang},
  journal={arXiv preprint arXiv:2403.15530},
  year={2024}
}

@ARTICLE{fei3dgssurvey,
  author={Fei, Ben and Xu, Jingyi and Zhang, Rui and Zhou, Qingyuan and Yang, Weidong and He, Ying},
  journal={IEEE Transactions on Visualization and Computer Graphics}, 
  title={3D Gaussian Splatting as New Era: A Survey}, 
  year={2024},
  volume={},
  number={},
  pages={},
}

@article{gaosurvey,
  author = {Wu, Tong and Yuan, Yu-Jie  and Zhang, Ling-Xiao and Yang, Jie and Cao, Yan-Pei and Yan, Ling-Qi and Gao, Lin},
  title = {Recent Advances in 3D Gaussian Splatting},
  journal = {Computational Visual Media},
  year = {2024},
  volume = {10}, 
  number = {},
  pages={613-642},
}

@misc{chen2025survey3dgaussiansplatting,
      title={A Survey on 3D Gaussian Splatting}, 
      author={Guikun Chen and Wenguan Wang},
      year={2025},
      journal={https://arxiv.org/abs/2401.03890}, 
}

@InProceedings{Kazhdan2006,
  author    = {Kazhdan, Michael and Bolitho, Matthew and Hoppe, Hugues},
  booktitle = {Proceedings of the Fourth Eurographics Symposium on Geometry Processing},
  title     = {Poisson Surface Reconstruction},
  year      = {2006},
  pages     = {61--70},
}

@Article{Kazhdan2013,
  author    = {Kazhdan, Michael and Hoppe, Hugues},
  journal   = {ACM Trans. Graph.},
  title     = {Screened Poisson Surface Reconstruction},
  year      = {2013},
  number    = {3},
  volume    = {32},
  articleno = {29},
  numpages  = {13},
}

@Article{Hou2022iPSR,
  author     = {Hou, Fei and Wang, Chiyu and Wang, Wencheng and Qin, Hong and Qian, Chen and He, Ying},
  journal    = {ACM Trans. Graph.},
  title      = {Iterative Poisson Surface Reconstruction (iPSR) for Unoriented Points},
  year       = {2022},
  number     = {4},
  volume     = {41},
  articleno  = {128},
  numpages   = {13},
}

@article{Liu2024DWG,
author = {Liu, Weizhou and Li, Jiaze and Chen, Xuhui and Hou, Fei and Xin, Shiqing and Wang, Xingce and Wu, Zhongke and Qian, Chen and He, Ying},
title = {Diffusing Winding Gradients ({DWG}): A Parallel and Scalable Method for 3D Reconstruction from Unoriented Point Clouds},
year = {2025},
journal = {ACM Trans. Graph.},
volume = {44},
number = {2},
articleno = {19},
numpages = {18},
}

@Article{Xu2023GCNO,
  author    = {Xu, Rui and Dou, Zhiyang and Wang, Ningna and Xin, Shiqing and Chen, Shuangmin and Jiang, Mingyan and Guo, Xiaohu and Wang, Wenping and Tu, Changhe},
  journal   = {ACM Trans. Graph.},
  title     = {Globally Consistent Normal Orientation for Point Clouds by Regularizing the Winding-Number Field},
  year      = {2023},
  number    = {4},
  volume    = {42},
  articleno = {111},
  numpages  = {15},
}

@Article{Lin2024WNNC,
  author={Lin, Siyou and Shi, Zuoqiang and Liu, Yebin},
  journal={ACM Trans. Graph.},
  title={Fast and Globally Consistent Normal Orientation based on the Winding Number Normal Consistency},
  year={2024},
  number={6},
  volume={43},
  articleno={189},
  numpages={19},
}

\end{document}